\documentclass[twoside]{article}

\usepackage[accepted]{include/aistats/aistats2025}
\usepackage[round]{natbib}

\usepackage[utf8]{inputenc} %
\usepackage[T1]{fontenc}    %
\usepackage{hyperref}       %
\usepackage{url}            %
\usepackage{booktabs}       %
\usepackage{multirow}
\usepackage{array}
\usepackage{multicol}
\usepackage{tabularx}
\usepackage{tabularray}
\UseTblrLibrary{booktabs}
\usepackage{algorithm}
\usepackage{algorithmic}

\usepackage{amsfonts}       %
\usepackage{nicefrac}       %
\usepackage{microtype}      %
\usepackage{xcolor}         %
\usepackage{listings}       %

\usepackage{enumitem}

\usepackage{graphicx}
\usepackage{subcaption}

\usepackage{include/preamble/titletoc}
\usepackage{amsmath}
\usepackage{amssymb}
\usepackage[utf8]{inputenc}
\usepackage{microtype}
\usepackage{graphicx}
\usepackage{booktabs} 
\usepackage{titletoc}
\usepackage{hyperref}

\usepackage{amsmath}
\usepackage{amssymb}
\usepackage{bbm} 
\usepackage{bm}
\usepackage{verbatim}
\usepackage{float}
\usepackage{color,soul}
\usepackage{enumitem}
\usepackage{mathtools}
\usepackage{hhline}
\usepackage[title]{appendix}
\usepackage[capitalise,noabbrev]{cleveref}
\usepackage{tabularx}
\usepackage{tikz}
\usepackage{wrapfig,booktabs}
\usepackage{xspace}
\usepackage{cancel}
\usepackage{sidecap}

\graphicspath{ {../figures/} }

\DeclarePairedDelimiterX{\infdivx}[2]{(}{)}{%
  #1\;\delimsize\|\;#2%
}

\crefname{appsec}{appendix}{appendices}
\Crefname{appsec}{Appendix}{Appendices}

\definecolor{mydarkblue}{rgb}{0,0.08,0.45}
\hypersetup{ %
    pdftitle={},
    pdfauthor={},
    pdfsubject={Proceedings of the International Conference on Machine Learning 2020},
    pdfkeywords={},
    pdfborder=0 0 0,
    pdfpagemode=UseNone,
    colorlinks=true,
    linkcolor=mydarkblue,
    citecolor=mydarkblue,
    filecolor=mydarkblue,
    urlcolor=mydarkblue,
    pdfview=FitH
}

\usetikzlibrary{shapes.geometric, arrows, bayesnet, calc, positioning}

\newcommand{\closer}[3]{{\kern-#1ex{#2}\kern-#3ex}}

\newcommand*{\disteq}{\overset{\text{d}}{=}}
\newcommand*{\iidsim}{\overset{\text{i.i.d.}}{\sim}}

\mathchardef\mhyphen="2D

\include{include/comments}

\newcommand{\pms}[1]{\ensuremath{{\scriptstyle\pm #1}}}
\newcommand{\emc}{E$\mathnormal{2}$MC}

\usepackage{xcolor}

\newcommand*{\iidsimnew}{\overset{\text{i.i.d.}}{\sim}}
\defcitealias{radical}{Learned-Miller and Fisher, 2003}

\definecolor{mydarkblue}{rgb}{0,0.08,0.45}
\hypersetup{ %
    pdftitle={},
    pdfauthor={},
    pdfsubject={},
    pdfkeywords={},
    pdfborder=0 0 0,
    pdfpagemode=UseNone,
    colorlinks=true,
    linkcolor=mydarkblue,
    citecolor=mydarkblue,
    filecolor=mydarkblue,
    urlcolor=mydarkblue,
    pdfview=FitH
}

\begin{document}

\runningtitle{Improving Pre-trained Self-Supervised Embeddings Through Effective Entropy Maximization}

\twocolumn[

\aistatstitle{Improving Pre-trained Self-Supervised Embeddings\\Through Effective Entropy Maximization}

\aistatsauthor{Deep Chakraborty \And Yann LeCun \And  Tim G. J. Rudner \And Erik Learned-Miller}

\aistatsaddress{UMass Amherst \And New York University \\ Meta - FAIR \And New York University \And UMass Amherst} ]

\begin{abstract}
A number of different architectures and loss functions have been applied to the problem of self-supervised learning (SSL), with the goal of developing embeddings that provide the best possible pre-training for as-yet-unknown, lightly supervised downstream tasks. 
One of these SSL criteria is to maximize the entropy of a set of embeddings in some compact space. 
But the goal of maximizing the embedding entropy often depends—whether explicitly or implicitly—upon high dimensional entropy estimates, which typically perform poorly in more than a few dimensions. 
In this paper, we motivate an effective entropy maximization criterion ({\emc}), defined in terms of easy-to-estimate, low-dimensional constraints. 
We demonstrate that using it to continue training an already-trained SSL model for only a handful of epochs leads to a consistent and, in some cases, significant improvement in downstream performance. 
We perform careful ablation studies to show that the improved performance is due to the proposed add-on criterion. 
We also show that continued pre-training with alternative criteria does not lead to notable improvements, and in some cases, even degrades performance.
\end{abstract}

\begin{figure*}[t]
    \centering
    \includegraphics[width=0.98\linewidth]{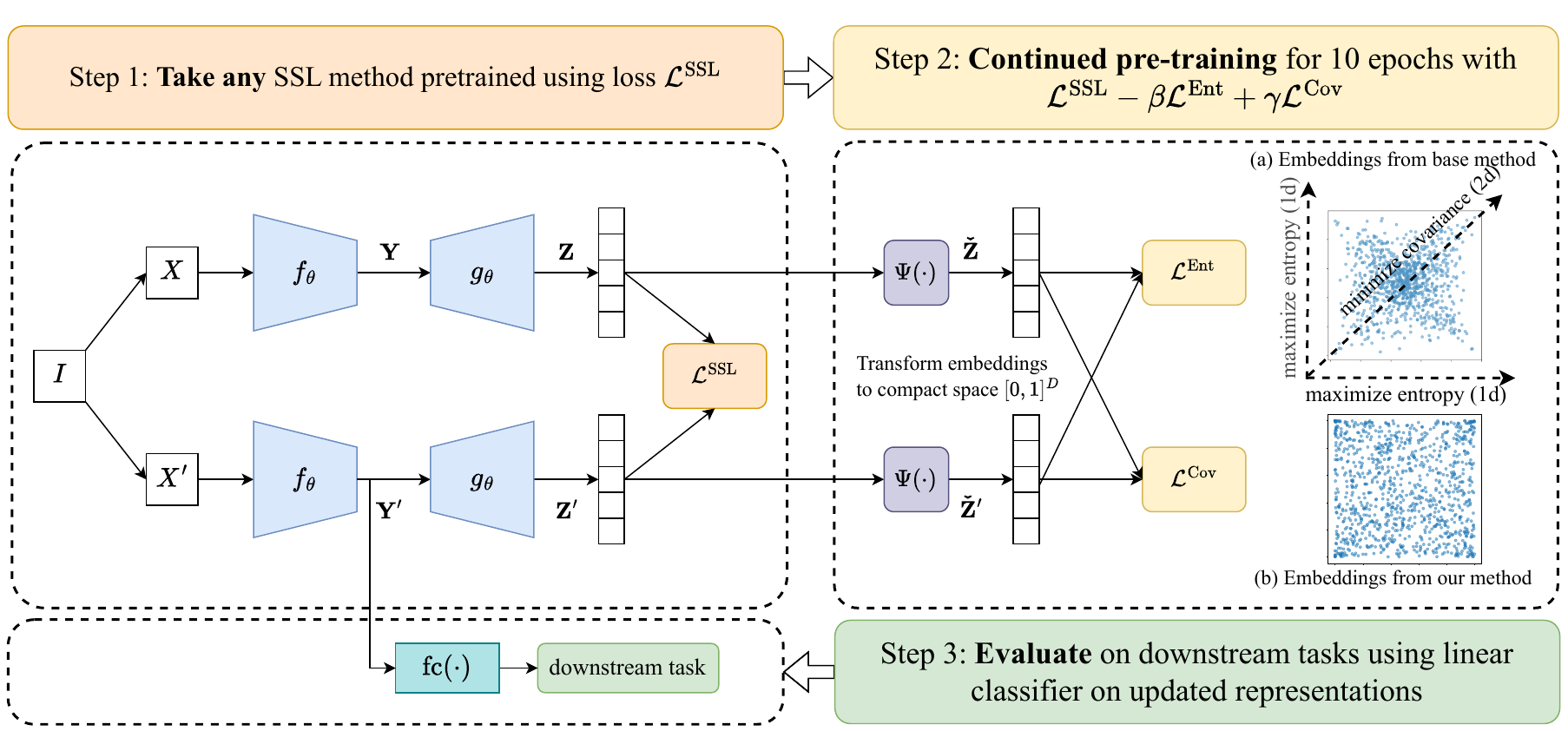}
    \caption{ An overview of our continued pre-training with {\emc} approach with three main stages: SSL model selection, training using augmented criterion, and evaluating updated representations on downstream tasks.}
    \label{fig:pipeline}
    \vspace*{-8pt}
\end{figure*}

\section{INTRODUCTION}
\label{sec:intro}

Self-supervised learning (SSL) methods are widely employed for pre-training features on unlabeled data and are highly effective for subsequent fine-tuning on a wide variety of downstream tasks \citep{simclr, byol, swav, vicreg}. 

In this paper, we ask whether it is possible to formulate a well-motivated, general-purpose criterion that allows {\it further improving} already-trained, highly-optimized SSL embeddings with only a handful of epochs of \mbox{{\it continued pre-training}}.

Like several previous works~\citep{nat,auh,mec,corinfomax}, we start with the principle of maximizing the entropy
of embeddings.  
One well-known motivation for this is that for a {\em discrete} embedding space, maximizing the entropy of a deterministic mapping preserves as much information as possible about the inputs. 
That is, such a maximum-entropy embedding maximizes the mutual information between the embedding and the input distribution \citep[see, for example,][]{deepinfomax}. 
Similar results hold for continuous embeddings under appropriate noise models
\citep[see, for example, discussion of the Gaussian channel in][]{cover1991elements}.

By maximizing the amount of information retained, one hopes to prepare as well as possible for future, as-yet-unknown, discrimination tasks. 
Our contribution is thus not the maximization of embedding entropy, but rather how we go about it.

\textbf{A fundamental problem with entropy maximization.} 
For any input distribution, a fixed neural network induces a distribution $p(z)$ on an embedding space. 
Since any neural network embedding is trained with a finite sample, we have no direct access to $p(z)$ and must attempt to maximize its entropy from a sample. 
Unfortunately, the amount of data required to get useful entropy estimates grows exponentially with the number of dimensions \citep{milimitation}. 
Practical estimators of joint entropy start to break down after just a handful ($\leq 10$) of intrinsic dimensions \citep{voronoi}. 
Thus, to claim that we are actually maximizing entropy in hundreds or thousands of dimensions is implausible. 
Instead, we focus on enforcing necessary, but not sufficient, conditions for maximum entropy. 

In particular, we choose conditions to enforce for which we have sufficient data: low-dimensional statistics.

These statistics are:\vspace*{-10pt}
\begin{enumerate}[leftmargin=20pt]
\setlength\itemsep{0pt}
    \item The one-dimensional entropy of each marginal component of our embeddings.
    \item The correlation of all pairs of marginals. 
\end{enumerate}
\vspace*{-8pt}
They have the following key properties:\vspace*{-10pt}
\begin{enumerate}[leftmargin=20pt]
\setlength\itemsep{0pt}
    \item They are {\em necessary} prerequisites for a maximum entropy joint distribution.
    \item We have plenty of data to estimate them, due to their low dimensionality. 
\end{enumerate}
\vspace*{-8pt}
At this point, we restate the fact that these statistics alone are {\em not sufficient} to enforce maximum entropy of a joint distribution. 
It is well known that joint distributions that are decorrelated and have maximum entropy marginals can have higher order (3rd order and higher) dependencies, dramatically reducing their joint entropy.

Surprisingly, we find that---without explicitly enforcing higher-dimensional constraints in our criterion---higher-order marginals of our embeddings naturally tend towards uniformity, resulting in practically useful embeddings. 
We demonstrate how this criterion can be added-on to \emph{any} pre-existing already-trained SSL model, which when further trained (\textit{continued pre-training}) for a handful of epochs (as few as ten), leads to consistent improvements in downstream classification tasks. 
In a resource-constrained compute environment, where a necessary downstream application is label deficient, gains from our proposed modifications are particularly higher, and can be used to leverage the full potential of powerful off-the-shelf SSL models by rapidly adapting their embeddings. Our code is available at \url{https://github.com/deepc94/e2mc}.

The main contributions of this paper are as follows:\vspace*{-8pt}
\begin{enumerate}[leftmargin=20pt]
\setlength\itemsep{0pt}
    \item
    We consider the problem of further improving already-trained, highly-optimized SSL embeddings using only limited computational resources.
    \item
    We motivate an {\it effective entropy maximization criterion} ({\emc}) grounded in information-theoretic principles and show that it can be used as an add-on criterion for popular SSL methods.
    \item
    We perform an empirical evaluation and find that with only a handful of epochs of continued pre-training under the proposed criterion, we achieve consistent and, in some cases, significant improvements in downstream-task performance across a selection of computer vision tasks.
\end{enumerate}
\section{BACKGROUND}
\label{sec:background}

\subsection{Joint-embedding self-supervised learning}
\label{back:ssl}
We refer to SSL methods that bring two `similar' input views (say, translated versions of the same image) closer together in the representation space while spreading apart different images, either explicitly \citep{simclr} or implicitly \citep{byol}, as {\em joint embedding methods}. 
These methods typically use Siamese style neural networks \citep{siamese} $f_\theta$ (encoder) to compute representation vectors $Y = f_\theta(X)$ and $Y' = f_\theta(X')$, where $X, X'$ are the two input image views. 
These representation vectors are then further transformed by an MLP $g_{\theta}$ (projector) to produce the final embeddings $Z_{\theta} = g_{\theta}(Y)$ and $Z'_{\theta} = g_{\theta}(Y')$. 
$Z, Z'$ are then optionally normalized (e.g., on the surface of a hypersphere) and used to compute one or more SSL loss functions $\mathcal{L}^{\textrm{SSL}}(\theta)$ (See `Step 1' in \Cref{fig:pipeline}). 
Once training is complete, the projector is discarded, and the representation vector $Y$ is used for downstream tasks.

Most methods employ regularization of some sort on the $Z$ embeddings in order to prevent trivial solutions or enforce desirable properties, or both. 
We have a similar goal wherein we take any such pre-trained SSL model and update its embeddings $Z$ by pre-training it for few additional epochs using our {\emc} approach (c.f. Step 2 in \Cref{fig:pipeline}).
Below, we briefly review some popular SSL methods, which we later improve using our proposed criterion.
\subsubsection{Variance-Invariance-Covariance Regularization (VICReg)}
\label{back:vicreg}

VICReg \citep{vicreg} is a feature decorrelation-based SSL method composed of the following:
\vspace*{-10pt}
\begin{enumerate}[label=(\alph*), leftmargin=20pt]
\setlength\itemsep{0pt}
    \item \textit{Invariance}: minimizes the euclidean distance between the embeddings of the original images and their augmented views $Z, Z'$, to learn features that remain consistent through input transformations. 
    \item \textit{Regularization}: consists of a \textit{variance preservation} term, that prevents the embedding components $Z_j$ from collapsing to a constant, and a \textit{covariance minimization} term that prevents redundant information from being encoded between any pair of embedding components $Z_j$ and $Z_k$ ($j\neq k$).
\end{enumerate}
\vspace*{-3pt}
The resulting loss function is defined as
\begin{align}
    \begin{split}
        \mathcal{L}^{\textrm{SSL}}(\theta)
        &=
        \frac{\lambda}{n} \left\|Z_{\theta} - Z'_{\theta}\right\|_2^2 
        + 
        \frac{\nu}{nd} \left\|K_{\theta}-\operatorname{diag}(K_{\theta})\right\|_F^2 \\
        &
        \quad
        +
        \frac{\mu}{d} \operatorname{Tr}\biggl(\operatorname{max}\Bigl(0, \eta - \sqrt{\operatorname{diag}(K_{\theta}) + \epsilon}\Bigr)\biggr),
    \end{split}\label{eq:vicregloss}
\end{align}
where $||\cdot||_{F}$ is the Frobenius norm, and we defined $K_{\theta} = \bar{Z}^\top_{\theta} \bar{Z}_{\theta}$, where $\bar{Z}_{\theta} = Z_{\theta} - \frac{1}{n} \sum\nolimits_{i = 1}^{n} Z_{\theta}^{i}$. $\eta$ is the target variance, and $\lambda$, $\nu$, and $\mu$ are the coefficients for the invariance, covariance, and variance terms respectively. 
The variance and covariance terms are computed symmetrically from both views $Z$ and $Z'$, and averaged.

\subsubsection{Swapped Assignments between Multiple Views (SwAV)}
SwAV \citep{swav} is an online-clustering based SSL method with the following features:
\vspace*{-10pt}
\begin{enumerate}[label=(\alph*), leftmargin=20pt]
\setlength\itemsep{0pt}
    \item \emph{Swapped prediction}: minimizes the cross-entropy between the cluster assignment $q_k$ of one augmented view, and the cluster prediction $p_k'$ using the other augmented view, to ensure consistent mapping of all views to the same cluster.
    \item \emph{Online clustering}: computes cluster centroids $c_k$ for $k$ clusters, and optimal cluster assignments $q_k$ (preventing collapse) on the fly.
    \item \emph{Multi-crop}: uses more than two views (usually lower resolution crops), to improve performance.
\end{enumerate}
\vspace*{-10pt}
The loss function used is
\begin{align}
        &
        \hspace*{50pt}
        \mathcal{L}^{\textrm{SSL}}(\theta)
        =
        -\sum\nolimits_{k} q_{\theta k} \log p_{\theta k}'\ , \label{eq:swavloss}\\
        &\textrm{where} \quad 
        p_{\theta k}' 
        = 
        \frac{\exp\left(\frac{1}{\tau} {\tilde{z}_\theta}^{'\top} c_{\theta k}\right)}{\sum\nolimits_{k'} \exp\left(\frac{1}{\tau} {\tilde{z}_\theta}^{'\top} c_{\theta k'}\right)}
        \quad \textrm{and}\quad 
        \tilde{z}_\theta' = \frac{z_\theta'}{\left\| z_\theta'\right\|_2},
        \nonumber
\end{align}
where $\tau$ is a temperature parameter, and the loss is computed over all data cases and augmentations.

\subsubsection{Simple Siamese Representation Learning (SimSiam)}
SimSiam \citep{simsiam} is a feature distillation based SSL method with the following features:
\vspace*{-10pt}
\begin{enumerate}[label=(\alph*), leftmargin=20pt]
\setlength\itemsep{0pt}
    \item \emph{Similarity}: maximizes the cosine similarity between the embedding of one augmented view and the \textit{predicted} embedding from the other view. 
    \item \emph{Asymmetry}: uses asymmetric Siamese branches with a predictor on one, and a \textit{stopgrad} on the other, to allow gradient flow through only one branch at a time.
\end{enumerate}
\vspace*{-10pt}
The loss function used is
\begin{align}
    \mathcal{L}^{\textrm{SSL}}(\theta)
    = 
    - \frac{p_\theta (Z_\theta)}{\left\| p_\theta (Z_\theta)\right\|_2} \cdot \frac{Z'_\theta}{\left\| Z'_\theta\right\|_2}\ , 
\end{align}
where $p_\theta$ is a MLP that predicts the embedding $Z'$ from $Z$, and $\left\| \cdot \right\|_2$ is the $l2$-norm. The loss is computed symmetrically for both views $Z$ and $Z'$.
\vspace*{-5pt}

\begin{figure*}[t]
\vspace*{-5pt}
    \centering
    \resizebox{\linewidth}{!}{%
    \begin{tabular}{cccc}
         \includegraphics[scale=0.25]{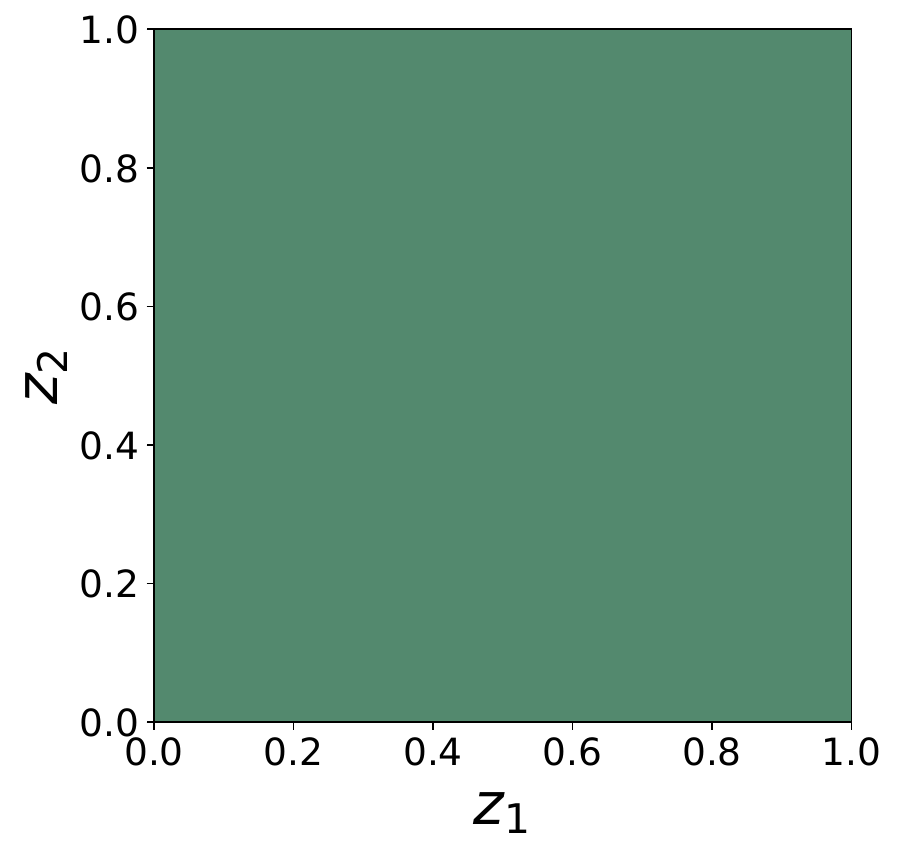}
         & 
         \includegraphics[scale=0.25]{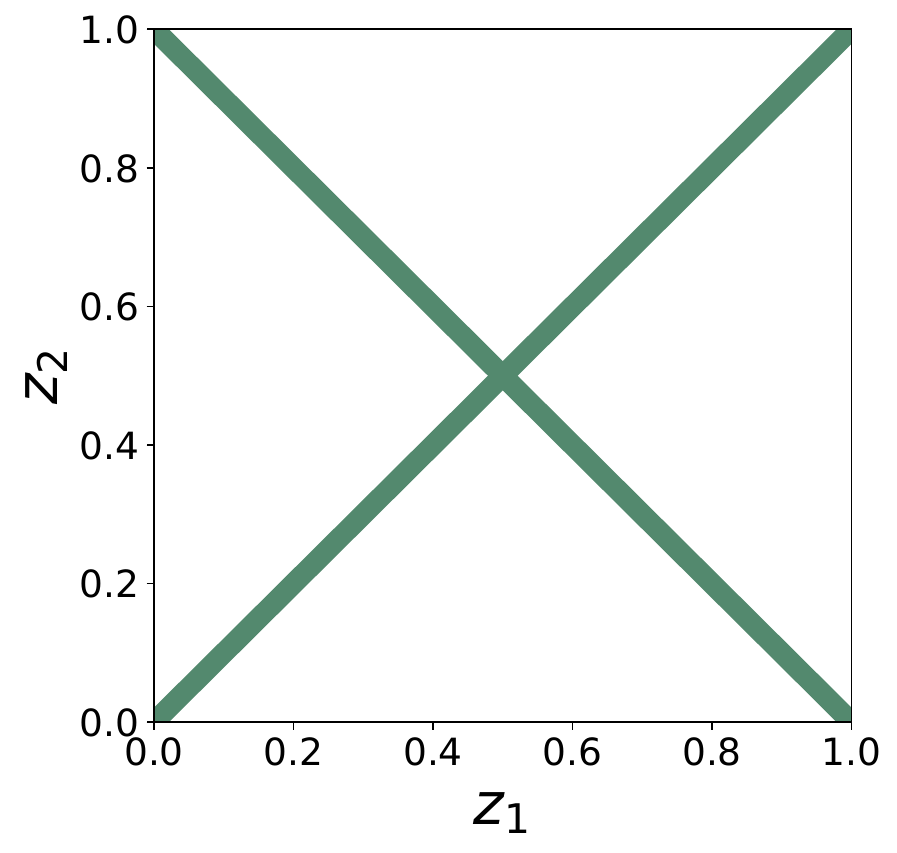}
         & 
         \includegraphics[scale=0.25]{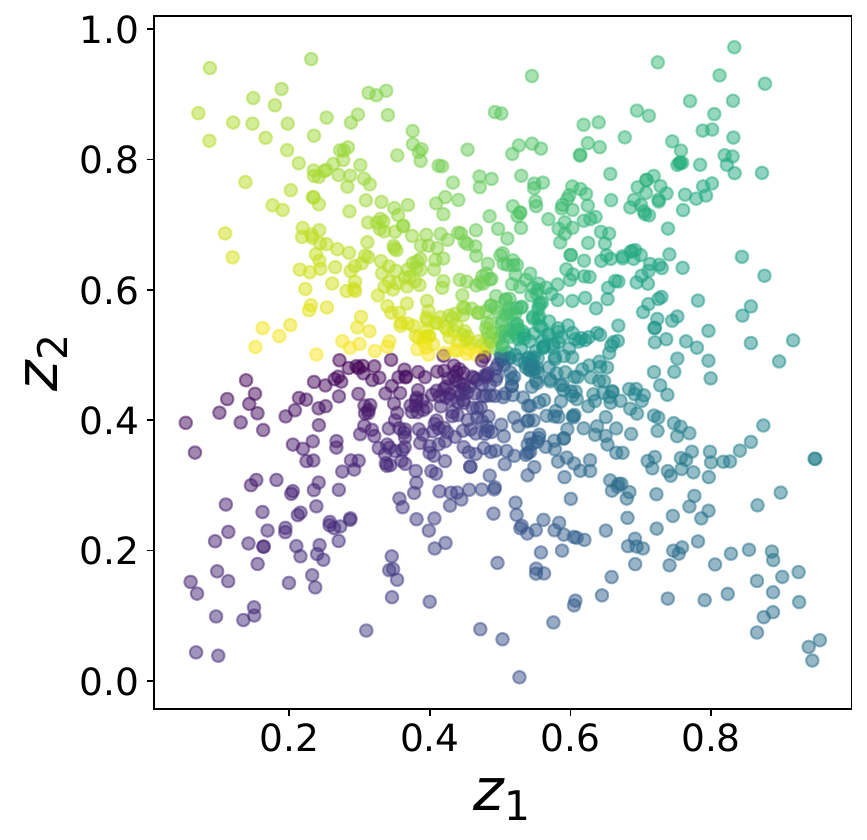}
         &
         \includegraphics[scale=0.25]{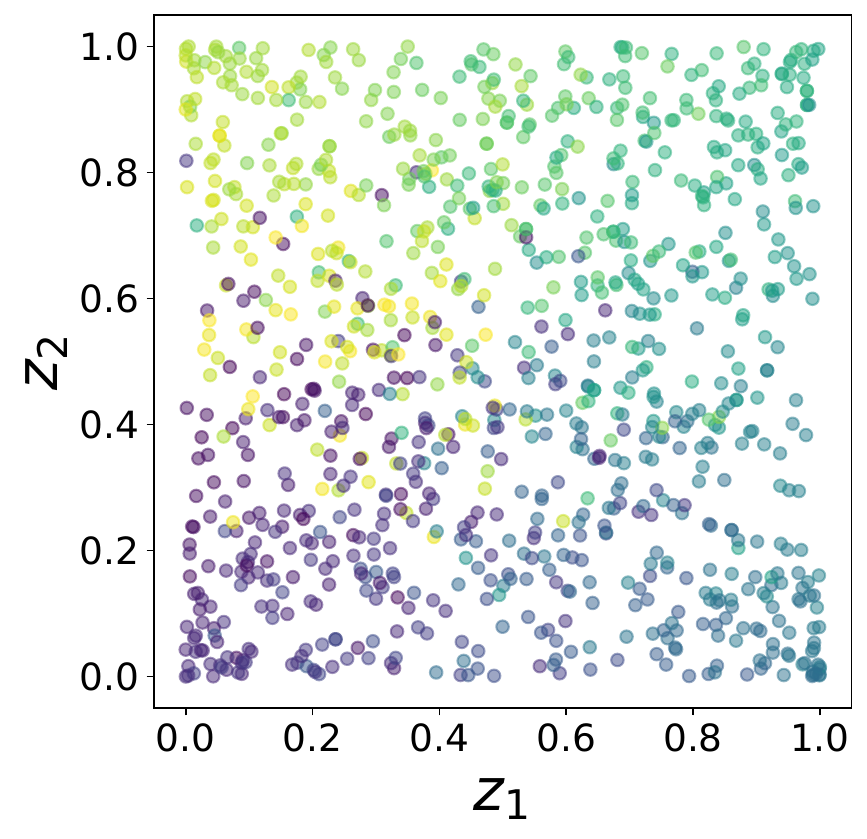} \\
         \quad (a) & \quad (b) & \quad (c) & \quad (d)
    \end{tabular}}
    \vspace*{-3pt}
    \caption{
        \textbf{(a)}. A 2-$d$ uniform distribution. 
        \textbf{(b)} An ``X'' distribution. Both (analytic) distributions have uniform (max-entropy) marginals and decorrelated components, and minimize our loss function. 
        \textbf{(c)} Example 2-d marginal distribution over a random pair from VICReg~\citep{vicreg} (after transformation to compact space).
        \textbf{(d)}~Our embeddings over the same pair of dimensions, where empirical results show, to our surprise, distributions with uniform 2-$d$ marginals despite the fact that this is not {\it explicitly} enforced by our loss. The colors denote the relative positions of actual points in embedding space \textit{before} (c) and \textit{after} (d) the application of our maximum-entropy criterion demonstrating how they are spread out by our method.
    }
    \label{fig:dists}
    \vspace*{-5pt}
\end{figure*}

\subsection{Differential entropy estimation of one-dimensional random variables}
\label{back:ent}

A key ingredient of our method is the entropy estimation of the one-dimensional marginals of our embeddings. Differential entropy estimation of one-dimensional distributions from a sample has a variety of solutions \citep[see, for example,][]{beirlant}. In this work, we use the $m$-spacings entropy \mbox{estimate \citep{Vasicek}}.
This estimator is statistically efficient, easy to compute, differentiable, and has been successfully applied to the independent components analysis (ICA) problem \citepalias{radical}. 
Though self-supervised embeddings are typically high dimensional, we will later show how we can leverage this one-dimensional estimator in our maximum-entropy criterion. 

The $m$-spacings estimator of a one-dimensional distribution's differential entropy is defined as 
\begin{align}
        & \widehat{\mathcal{H}}_{j}(\check{Z}^{1}, ..., \check{Z}^{n})
        \\
        &
        ~
        =
        \frac{1}{(n-m)} \sum\nolimits_{i=1}^{n-m} \log \left(\frac{n+1}{m}\left(\check{Z}_{j}^{(i+m)} - \check{Z}_{j}^{(i)}\right)\right)
        \nonumber
\end{align}
for the $j$th dimension of $\check{Z}$, where $\check{Z} \in [0, 1]^D$ is the compact version of the embedding $Z$ (details in Sec.~\ref{sec:compact}). 
Parenthetical superscripts indicate the position in the ordering $\check{Z}_{j}^{(1)} \leq \check{Z}_{j}^{(2)} \leq \cdots \leq \check{Z}_{j}^{(n)}$, and $\check{Z}_{j}^{(i+m)} - \check{Z}_{j}^{(i)}$ is known as a spacing of order $m$ (typically $m=\sqrt{n}$).
\section{A MAXIMUM-ENTROPY AUGMENTATION CRITERION}
\label{sec:method}

In this section, we motivate a simple maximum-entropy augmentation criterion that can be used to improve already-trained SSL embeddings with only a handful of epochs of continued pre-training.

Unlike all the other methods of which we are aware, we focus only on properties of the one- and two-dimensional marginal distributions, and speculate that by focusing on properties that are more reliably estimated with moderate sample sizes, we might be able to obtain a more useful criterion.

To motivate an effective maximum-entropy criterion, we start with an observation that the following facts about distributions over the unit cube are mathematically equivalent~\citep{cover1991elements}:\vspace*{-5pt}
\begin{enumerate}[leftmargin=20pt]
\setlength\itemsep{0pt}
    \item The joint distribution has maximum joint entropy.
    \item The joint distribution is uniform.
    \item The one-dimensional marginal distributions are maximum entropy (i.e., uniform) {\it and} the components are mutually independent.
\end{enumerate}
We use the third characterization to design our loss function.
This characterization for formulating a self-supervised learning criterion requires (i) an effective approach to estimating the entropy of one-dimensional marginal distributions and (ii) a method for encouraging 
mutual independence.

To obtain a good estimate of the marginal entropies, we leverage the $m$-spacings estimator (c.f. \Cref{back:ent}).
Unfortunately, mutual independence of the components is a property of the joint distribution, and we believe that it is too high-dimensional to achieve directly. 
Instead, we consider a {\it necessary, but not sufficient, condition} for mutual independence: 
decorrelation of all pairs of embedding dimensions.
Criteria that serve this purpose have been used in both VICReg and other SSL methods to attempt to move embeddings towards independent features \citep{vicreg,vicregpairs} but not, to our knowledge, in conjunction with the idea of maximizing marginal entropies.

Unfortunately,  enforcing 
decorrelation of all pairs of embedding dimensions does not guarantee mutual statistical independence.
Hence, maximizing marginal entropies while decorrelating embedding dimensions is not sufficient to guarantee maximum entropy of the joint distribution.
We nevertheless press on and ask:
\begin{quote}
What kinds of distributions have maximal marginal entropy {\it and} are decorrelated but do not have maximum joint entropy?
\end{quote}
Consider \Cref{fig:dists}.
Part (a) shows a two-dimensional uniform distribution, which maximizes the joint entropy and minimizes our loss function.
Part (b) is what we call the ``X'' distribution, which also has uniform marginals and  diagonal covariance (i.e., no correlations between components).
In principle, either of these distributions could emerge under the criterion described above.
Part (c) shows a 2-d marginal of VICReg, which is clearly non-uniform. Surprisingly, our loss, which enforces uniformity only of 1-d marginals, also produces nearly uniform 2-d marginals as shown in (d), instead of alternatives like the ``X'' distribution. One possible explanation could be that the inductive bias of such deep networks might make it difficult to produce non-smooth distributions like the ``X'' distribution. 
\subsection{Specifying a maximum-entropy augmentation criterion}

In this section, we formalize the specific criterion from the discussion above.
To define this criterion, we first transform embedding samples $Z \in \mathbb{R}^d$ to lie in a compact space, 
and consider the transformed embedding random variable $\smash{\check{Z} \in [0,1]^d}$ instead, for applying our criterion. 
We defer the discussion of this transformation to \Cref{sec:compact}.
Finally, given an arbitrary SSL method pre-trained using loss function $\mathcal{L}^{\textrm{SSL}}(\theta)$, we define the constrained optimization problem
\begin{align}
    \min\nolimits_{\theta} \mathcal{L}^{\textrm{SSL}}(\theta)
    \qquad
    \textrm{subject~to}\quad
    \mathcal{L}^{\mathrm{Entropy}}(\theta)
    \geq
    C_{1}, &\notag
    \\
    \quad \textrm{and}\quad 
    \mathcal{L}^{\mathrm{Covariance}}(\theta)
    \leq
    C_{2}. &
\end{align}
In practice, we express this objective equivalently as
\begin{align}
    \mathcal{L(\theta)} 
    = 
    \mathcal{L}^{\mathrm{SSL}}(\theta) - \beta \mathcal{L}^{\mathrm{Entropy}}(\theta) + \gamma \mathcal{L}^{\mathrm{Covariance}}(\theta), \label{eq:fullloss}
\end{align}
where $\beta,\gamma \in \mathbb{R}$ are hyperparameters. For transformed embeddings $\check{Z}_{\theta}$ and $\check{Z}'_{\theta}$ of views $X$ and $X'$, we have
\begin{align}
        &
        \mathcal{L}^{\mathrm{Entropy}}(\theta)
        \\
        &
        ~
        =
        \frac{1}{d} \sum\nolimits_{j = 1}^{d} \Big( 
        \widehat{\mathcal{H}}_{j}(\check{Z}_{\theta}^{1}, \cdots, \check{Z}_{\theta}^{n})
        +
        \widehat{\mathcal{H}}_{j}(\check{Z}_{\theta}^{'1}, \cdots, \check{Z}_{\theta}^{'n}) \Big).
        \nonumber
    \label{eq:margent}
\end{align}
And, letting ${\check{\bar{Z}}}_{\theta} = \check{Z}_{\theta} - \frac{1}{n} \sum\nolimits_{i = 1}^{n} \check{Z}_{\theta}^{i}$ and ${\check{\bar{Z}}'_{\theta}} = \check{Z}'_{\theta} - \frac{1}{n} \sum\nolimits_{i = 1}^{n} \check{Z}_{\theta}^{'i}$ for $X$ and $X'$, we have
\begin{align}
    &
    \mathcal{L}^{\mathrm{Covariance}}(\theta)
    \\
    &
    ~
    =
    \frac{1}{nd} \big(\left\|(K_{\theta}-\operatorname{diag}(K_{\theta}))\right\|_F^2
    +
    \left\|(K'_{\theta}-\operatorname{diag}(K'_{\theta}))\right\|_F^2 \big) ,
    \nonumber
\end{align}
where $||\cdot||_{F}$ is the Frobenius norm, and we defined $K_{\theta} = \check{\bar{Z}}^\top_{\theta} \check{\bar{Z}}_{\theta}$ and $K'_{\theta} = \check{\bar{Z}}^{'\top}_{\theta} \check{\bar{Z}}'_{\theta}$.

We estimate the marginal entropies $\widehat{\mathcal{H}}_{j}$ for each embedding dimension $j$ using the $m$-spacings estimator (c.f. \Cref{back:ent}) and average them in the final loss.
We estimate the sample covariance for every pair of embedding dimensions $j,k$ and $j\neq k$ using the same estimator as VICReg (c.f. \Cref{back:vicreg}, \Cref{app:components}).

\pagebreak

Under this formulation, maximizing $\mathcal{L}^{\mathrm{Entropy}}(\theta)$ maximizes the marginal entropies, and minimizing $\mathcal{L}^{\mathrm{Covariance}}(\theta)$ corresponds to minimizing the squared off-diagonal entries of the sample covariance computed from the embedding.

\subsection{Transformation to a compact space}
\label{sec:compact}
Maximizing entropy on a non-compact space such as $\mathbb{R}^d$ is not meaningful, since the data can simply be spread out without bound. 
That is, our methods are meaningfully applied only on compact spaces. 
We discuss the maximization of entropy on two compact spaces: the unit hypercube and the surface of the unit hypersphere. 
We begin with the hypercube.

For SSL methods that produce embeddings in $\mathbb{R}^d$ and do not normalize their final embeddings (e.g., VICReg), we construct a transformation $\smash{{\Psi: \mathbb{R} \rightarrow [0,1]}}$, and apply it to every embedding component $\smash{\check{Z}_j = \Psi(Z_j)}$, such that the transformed embedding $\check{Z} = [\check{Z}_1, \cdots, \check{Z}_d]$ lies in a unit hypercube of $d$ dimensions, with an implicit joint distribution $\smash{p(\check{z}_1, ..., \check{z}_d)}$ over the hypercube. We simply let $\Psi$ be the sigmoid transformation, $\smash{\Psi(Z_j) = 1 / (1 + \exp(-Z_j))}$ and apply our loss function to this transformed embedding.  

However, methods that do normalize their final embeddings to be on the hypersphere (e.g., SwAV, SimSiam) present a unique challenge. In particular, if we produce uniform marginal embeddings and then normalize, the resulting distribution on the hypersphere will be far from uniform. In particular, mass will be much greater in directions corresponding to the corners of the hypercube, since the projections there will accumulate density along the longer diagonal directions of the hypercube.  
How then can we construct $\Psi$ such that maximizing the entropy of the compact embeddings $\check{Z}$ translates to a uniform distribution on the hypersphere
when the original embeddings $Z$ are normalized?

To answer this question, we use a simple result that is often used to draw samples uniformly from the surface of a hypersphere \citep{spheretrick}: 
If we construct an embedding vector $Z$ whose components $Z_j$ are independent zero-mean, unit-variance Gaussians $Z_j \iidsimnew \mathcal{N}(0, 1)$, 
then the normalized embedding vector $\tilde{Z} = Z / \left\| Z \right\|_2$ maps uniformly onto the surface of the unit hypersphere $\mathcal{S}^{d-1}$. In practice, 
we apply this result by letting $\Psi$ be the cumulative density function (CDF) of the zero-mean, unit-variance Gaussian, $\smash{\Psi(Z_j) = 0.5 (1 + \textrm{erf}(x/\sqrt{2}))}$, and then applying our entropy maximization criterion to the transformed embeddings to produce a uniform distribution. 
This is possible due to the \emph{probability integral transform}, in  which a continuous random variable is mapped through its own CDF to become uniformly distributed. 
This implies that the distribution over transformed variables $\smash{p(\check{z}_j) \disteq \mathcal{U}[0, 1]}$ if and only if the distribution over original embeddings $\smash{p(z_j) \disteq \mathcal{N}(0, 1)}$. Our criterion thus ensures that the components of the embedding distribution prior to normalization are normal with zero mean and unit variance. In addition, our term to minimize correlation helps to minimize dependencies among the unit variance marginals. 

Using these two methods of transforming to compact spaces, our criterion can be applied to SSL methods irrespective of their normalization strategy.

\setlength{\tabcolsep}{5.7pt}
\begin{table*}[t!]
\centering
\small
\caption{
    \textbf{Evaluation on ImageNet.} We report Top-1-Accuracy (\%) on ImageNet validation set using classifiers trained on SSL embeddings, before and after continued pre-training. Best result in each category is marked bold if a clear winner exists, along with standard errors over three random trials. We run statistical significance tests in \Cref{app:statsig}. Results marked with $^\dagger$ are taken from the original papers, and~$^*$ are reproduced results that differ from reported results in the original paper despite best attempts.
    Note: No numbers were reported for SimSiam in the semi-supervised setting, and hence omitted. \label{tab:results}
}
\vspace*{-8pt}
\resizebox{\linewidth}{!}{%
\begin{tabular}{@{}lrcccccc@{}}
\toprule
                            &           & \multicolumn{3}{c}{Linear Evaluation}                              & \multicolumn{1}{l}{} & \multicolumn{2}{c}{Semi-supervised learning}                        \\ \cmidrule(lr){3-5} \cmidrule(l){7-8}
Method                      & Epoch        & 1\% labels         & 10\% labels        & 100\% labels                &           & 1\% labels                      & 10\% labels                       \\ \midrule
VICReg base \citep{vicreg}  & {1,000}   & 53.50 $\pms{0.11}$ & 66.57 $\pms{0.02}$ & \hspace{-2em}73.20$^\dagger$             &           & 54.53$^* \pms{0.12}$             & 67.97$^* \pms{ 0.03}$             \\
VICReg continued                & {1,010}   & 53.51 $\pms{0.07}$ & 66.57 $\pms{0.06}$ & 73.16 $\pms{0.02}$          &           & --                           & --                           \\
VICReg + {\emc} (\bf ours)  & {1,010}   & \hspace{0.4em}\textbf{54.54} $\pms{0.05}$ & \hspace{0.3em}\textbf{66.82} $\pms{0.05}$ & \hspace{0.4em}\textbf{73.45} $\pms{0.07}$          &        & \hspace{0.3em}\textbf{55.05} $\pms{ 0.08}$                      & \hspace{0.3em}\textbf{68.12} $\pms{ 0.04}$                     \\ \midrule
SwAV base \citep{swav}      & 400     & 52.34 $\pms{0.07}$ & 67.61 $\pms{0.02}$ & \hspace{-2em}74.30$^\dagger$             &           & 52.57 $\pms{ 0.15}$                       & 69.25 $\pms{ 0.05}$                      \\
SwAV continued                  & 410     & 52.31 $\pms{0.07}$ & 67.56 $\pms{0.05}$ & 74.31 $\pms{0.02}$          &           & --                           & --                           \\
SwAV + {\emc} (\bf ours)    & 410     & \hspace{0.4em}\textbf{53.40} $\pms{0.01}$ & \hspace{0.3em}\textbf{67.73} $\pms{0.03}$ & \hspace{0.4em}\textbf{74.44} $\pms{0.03}$          &           & 52.70 $\pms{ 0.54}$                    & 69.24 $\pms{ 0.02}$                       \\ \midrule
SwAV base  \citep{swav}     & 800     & 53.70 $\pms{0.05}$ & 68.86 $\pms{0.03}$ & \hspace{-2em}75.30$^\dagger$             &           & 53.89$^\dagger \pms{ 0.13}$ & 70.22$^\dagger \pms{ 0.05}$ \\
SwAV continued                  & 810     & 53.69 $\pms{0.05}$ & 68.87 $\pms{0.04}$ & 75.32 $\pms{0.01}$          &           & --                           & --                           \\
SwAV + {\emc}  (\bf ours)   & 810     & \hspace{0.4em}\textbf{55.27} $\pms{0.07}$ & \hspace{0.3em}\textbf{68.98} $\pms{0.02}$ & \hspace{0.4em}\textbf{75.41} $\pms{0.02}$          &        & 53.94 $\pms{ 0.30}$                      & 70.32 $\pms{ 0.05}$            \\ \midrule
SimSiam base \citep{simsiam}& 100     & 43.71 $\pms{0.04}$ & 60.15 $\pms{0.02}$ & \hspace{-1.9em}68.37$^*$ &           & --                      & --                       \\
SimSiam continued               & 110     & 43.78 $\pms{0.05}$ & 60.23 $\pms{0.08}$ & 68.45 $\pms{0.08}$          &           & --                           & --                           \\
SimSiam + {\emc}  (\bf ours)& 110     & 43.78 $\pms{0.06}$ & 60.23 $\pms{0.07}$ & \hspace{0.3em}\textbf{68.52} $\pms{0.05}$          & \textbf{} & --                      & --                     \\ \bottomrule
\end{tabular} }
\end{table*}

\section{RELATED WORK}
\label{sec:related_work}

Here we mainly review prior approaches to improving self-supervised embeddings using explicit information maximization objectives, and how they relate to ours. 
{\bf Log determinant maximization.}  \citet{corinfomax} propose \emph{CorInfoMax}, that maximizes mutual information between similar views in latent space, while preventing dimensional collapse by spreading embeddings in this space. 
They use the log-determinant mutual information as a second-order approximation of the mutual information, and maximize the log-determinant of the covariance matrix as a measure of the spread of latent vectors (under a Gaussian model assumption). 
\mbox{\citet{mec}} also posit that the most generalizable embeddings should have the maximum possible entropy in order to avoid bias from pretext tasks.
They maximize a related objective---minimum coding length---as a computationally tractable surrogate for entropy of high dimensional embeddings, and offer a simplified version that is less computationally demanding, but suffers from the same distributional assumptions as before.
\citet{infovicreg} show that, under the assumption that the input data is a mixture of Gaussians, VICReg maximizes an upper bound to the embedding entropy, through an approximation of the log determinant of the covariance matrix.
We show that our criterion outperforms these methods without making distributional assumptions while also relying solely on first and second order statistics.

{\bf Manifold capacity.} In another recent method, \emph{Maximum Manifold Capacity Representations}, \citet{mmcr} proposed that image representations should lie on compact sub-manifolds in the representation space that are well separated from each other. 
Their method favors low-rank manifolds constructed by averaging embeddings from multiple views of an image, and then maximizing the nuclear norm of the embedding matrix 
rather than the log-determinant of the covariance matrix like the previous approaches.
\citet{infommcr} show that this method also maximizes a lower bound on the mutual information between input views.
In practice, we find that this method cannot be used as an add-on criteria in the continued pretraining setting like ours, lowering the performance of base SSL methods. 
{\bf Noise as targets.} 
In \textit{Noise as Targets} (NAT), \mbox{\citet{nat}} propose to map input samples to a fixed set of embeddings uniformly sampled from the surface of a hypersphere. 
Though this approach has similarities with our objective, an important difference is that NAT strives to match an empirical {\it sample} from a uniform distribution, 
whereas our loss function enforces properties of a true uniform distribution.
In  {\it Alignment and Uniformity on the \mbox{Hypersphere}} (AUH), \citet{auh} propose a way to distribute embeddings by minimizing the energy configuration of points using pairwise Gaussian potentials.
They show that the only distribution from which their objective admits samples, in the limit, is the uniform distribution. \citet{mmd} generalize their method further by minimizing the maximum mean discrepancy (MMD) between the embedding distribution and uniform distribution using rotation invariant kernels instead of the RBF kernel.

Our {\it maximum-entropy} criterion ({\emc}) also mimics certain properties of a fully uniform distribution on a compact embedding space.
The key difference between these methods and ours is that we enforce properties of the one- and two-dimensional marginals of the distribution, rather than operating directly on properties of the joint distribution.
We also find in our experiments that {AUH} fails to produce improvements as large as ours in the challenging continued pretraining setting. 
We hypothesize that since AUH evaluates energy potentials of the high-dimensional joint distribution rather than of single-dimensional marginal distributions, it is less sample efficient. 
Our method also does not rely on training from scratch unlike all of the above methods, and is able to adapt existing SOTA embeddings for better downstream performance. 
Works like \mbox{EMP-SSL} \citep{empssl} show this is increasingly important in the realm of extant resource intensive SSL methods.
{\bf High-dimensional estimators of differential entropy.}
Recent advances have also been made in developing high-dimensional estimators of differential entropy for regularizing neural network training \citep{knife, remedi}.
While these methods propose an upper bound entropy estimator, it is still subject to the data inefficiency of plug-in estimators in even moderately high dimensions \citep{optimality}. 
It is unclear therefore, how their approach would scale to dimensions as large as 8,192 (the highest dimensionality addressed in our paper). 
These methods also require fitting a neural network with additional parameters and training with large batch sizes, whereas ours does not.
\pagebreak

\section{EMPIRICAL EVALUATION}
\label{sec:experiments}

Our experimental setup uses a three-stage approach (c.f. \Cref{fig:pipeline} for an overview):\vspace*{-8pt}
\begin{enumerate}[leftmargin=20pt]
\setlength\itemsep{-2pt}
    \item
    \textbf{Selecting} a base SSL method with publicly available checkpoints,
    \item
    \textbf{Continued pre-training} on the base dataset using the base SSL method augmented with our criterion, and 
    \item
    \textbf{Evaluating} the representations learned by the backbone network using classifiers trained on downstream datasets. 
\end{enumerate}
\vspace*{-8pt}

While our approach can be used with \emph{any} joint-embedding SSL method, we focus on three popular methods---VICReg, SwAV, and SimSiam---to demonstrate the versatility of our criterion.
These methods do not require negative examples, work well with small batch sizes, and have official checkpoints and code that can be modified to incorporate our criterion. 

In the continued pre-training stage, we train on the same dataset that was used for pre-training the base SSL method (ImageNet; \citealp[]{imagenet}), but with a fixed reduced learning rate 
and batch size. 
We train for \textit{exactly} 10 additional epochs using the prescribed criterion, and report results using this updated model.
All other hyperparameters associated with the base SSL method, including data augmentations, optimizers, and loss coefficients, are kept identical.
This allows us to treat the base SSL method as a black box, and leaves us with only a few hyperparameters to tune for our method, namely the coefficients $\beta$ and $\gamma$ for our proposed loss in \Cref{eq:fullloss}. 
See \Cref{app:implement} for implementation details and pseudocode (\Cref{alg:method}).

\subsection{Evaluation on ImageNet}

We evaluate methods on ImageNet \citep{imagenet} using the final representations (2048-d) from the ResNet-50 backbone \citep{resnet} in two ways: (i)~\emph{linear} evaluation by training a classifier on top of the frozen backbone, and (ii)~\emph{semi-supervised} evaluation by finetuning the whole backbone with a classifier on a subset of available labels. 
\Cref{tab:results} shows the top-1 accuracy on the ImageNet validation set of classifiers trained using different subsets of available labels (predefined by \citealp[]{simclr}).
In the linear evaluation scenario, our method ``[SSL]+{\emc}'' surpasses almost all baselines from just 10 epochs of continued pre-training, with larger gains in label deficient (1\% subset) settings.
For SwAV, the maximum-entropy \emph{updated} 400-epoch model in the 1\% setting has comparable performance with the 800-epoch base model (53.4 vs 53.7 respectively), which suggests our method has fast convergence obviating the need for longer pre-training.
In the semi-supervised learning scenario, our method outperforms the VICReg baseline, and is on par with SwAV. Note that this setting is more challenging to show improvements on, as fine-tuning the backbone can change the final embeddings significantly, and subvert any benefits from learning maximum-entropy embeddings.
On SimSiam, our method shows little improvement over the baseline, which could be because the SimSiam checkpoint is {\it less} trained compared to other methods (only 100 epochs), and benefits from our method may only emerge late in the pre-training phase where base models already have near-optimal performance.

\subsection{Transfer Learning on Other Datasets}

Following \cite{vissl} and \cite{pirl}, we show how our updated representations transfer to downstream tasks on other datasets. \Cref{tab:transfer} shows top-1 accuracy from linear classification on the challenging iNat18 dataset \citep{inaturalist} (with over 8000 fine-grained classes), and mAP for multi-label object classification on \mbox{VOC07~\citep{pascal}}. Again, we show consistent improvements for iNat18, and better or comparable performance to base models on VOC07.

\setlength{\tabcolsep}{8pt}
\begin{table}[t]
    \centering
    \small
    \captionof{table}{\textbf{Transfer Learning.} We report Top-1-Accuracy (\%) of linear classifier trained on iNat18, and mAP of linear SVM trained on VOC07 datasets. Best results shown in bold. Results marked with $^\dagger$ are taken from the original papers, and~$^*$ are reproduced results that differ from reported results in the original paper despite best attempts. \vspace*{-4pt}\label{tab:transfer}}
    \begin{tabular}{@{}lrlc@{}}
        \toprule
        Method          & Epoch      & iNat18             & VOC07           \\ \midrule
        VICReg base     & 1,000     & 47.00$^\dagger$    & \hspace{0.1em}86.60$^\dagger$ \\
        VICReg + {\emc} & 1,010 & \textbf{47.18} $\pms{0.11}$ & \textbf{86.80}           \\ \midrule
        SwAV base       & 400        & 46.00              & \textbf{88.38}           \\
        SwAV + {\emc}   & 410    & \textbf{46.71} $\pms{0.17}$ & \hspace{-0.2em}88.24           \\ \midrule
        SwAV base       & 800        & 49.08$^*$          & \hspace{0.2em}88.56$^*$       \\
        SwAV + {\emc}   & 810    & \textbf{49.72} $\pms{0.20}$ & \textbf{88.69}           \\ \midrule
        SimSiam base    & 100     & 38.75              & \textbf{84.62}           \\
        SimSiam + {\emc} & 110 & \textbf{38.99} $\pms{0.20}$ & \hspace{-0.2em}84.54           \\ \bottomrule
    \end{tabular}%
\end{table}

\subsection{Ablation Study: Effect of different continued pre-training criteria}

To disambiguate performance gains of continued pre-training using our method from continued pre-training using other criteria, we train the base SSL method for the same number of epochs as our method, but with the other criteria.

One important setting is to simply train the base method longer using only its original loss function(s) (``[SSL] continued''). 
As seen in \Cref{tab:results}, continued training with base criteria produces no additional gains for any of the SSL methods under consideration.

Another important setting is where we replace our maximum-entropy criteria with similar criteria proposed by others. 
\Cref{tab:ablation} shows the results of doing continued pre-training on SwAV with other criteria. 
We chose the SwAV model for these experiments as it has several features that are amenable to the other criteria we experiment with. 
VCReg \citep{vicreg} uses covariance minimization similar to our criterion, but variance regularization produces inferior gains compared to entropy maximization. 
The uniformity loss proposed in AUH \citep{auh} also fails to produce improvements as large as our method, or even VCReg, and we hypothesize that their method is less sample efficient because it optimizes properties of the high dimensional joint distribution, instead of lower dimensional criteria such as ours. 
Finally, MMCR \citep{mmcr} uses an alternative information criterion compared to these methods, and we show that it degrades the performance of the base model instead of improving it. This suggests that not all information criteria are created equal, and some such as MMCR, though good standalone methods, do not play as nicely with popular SSL criteria. 

\setlength{\tabcolsep}{2pt}
\begin{table}[t]
    \centering
    \small
    \captionof{table}{\textbf{Effect of different continued pre-training criteria.} We report Top-1-Accuracy (\%) of linear classifier trained on ImageNet, using SwAV base (800 epochs) \citep{swav} and SwAV + \textit{other criteria} (810 epochs) including $^{(a)}$~\cite{vicreg}, $^{(b)}$~\cite{mmcr}, $^{(c)}$~\cite{auh}, and $^{(d)}$~ours. Best results are shown in bold, and second best are underlined. Details of these alternative criteria can be found in \Cref{app:implement}. \vspace*{-4pt} \label{tab:ablation}}
    \begin{tabular}{@{}lccc@{}}
        \toprule
        Method        & 1\% labels         & 10\% labels        & 100\% labels       \\ \midrule
        SwAV base     & 53.70 $\pms{0.05}$ & 68.86 $\pms{0.03}$ & \hspace{-2.0em}75.30$^\dagger$    \\
        SwAV continued   & 53.69 $\pms{0.05}$ & 68.87 $\pms{0.04}$ & 75.32 $\pms{0.01}$ \\
        SwAV + VCReg$^{(a)}$  & \underline{54.02} $\pms{0.05}$ & \underline{68.88} $\pms{0.03}$ & \underline{75.36} $\pms{0.02}$ \\
        SwAV + MMCR$^{(b)}$   & 53.30 $\pms{0.02}$ & 68.77 $\pms{0.04}$ & 75.27 $\pms{0.01}$ \\
        SwAV + AUH$^{(c)}$   & 53.84 $\pms{0.07}$ & 68.90 $\pms{0.04}$ & 75.33 $\pms{0.01}$ \\
        SwAV + {\emc}$^{(d)}$ & \textbf{55.27} $\pms{0.07}$ & \textbf{68.98} $\pms{0.02}$ & \textbf{75.41} $\pms{0.02}$ \\ \bottomrule 
    \end{tabular}%
\end{table}

\begin{figure*}[t]
    \centering
    \begin{minipage}[t]{0.4\linewidth}
          \includegraphics[width=0.95\linewidth]{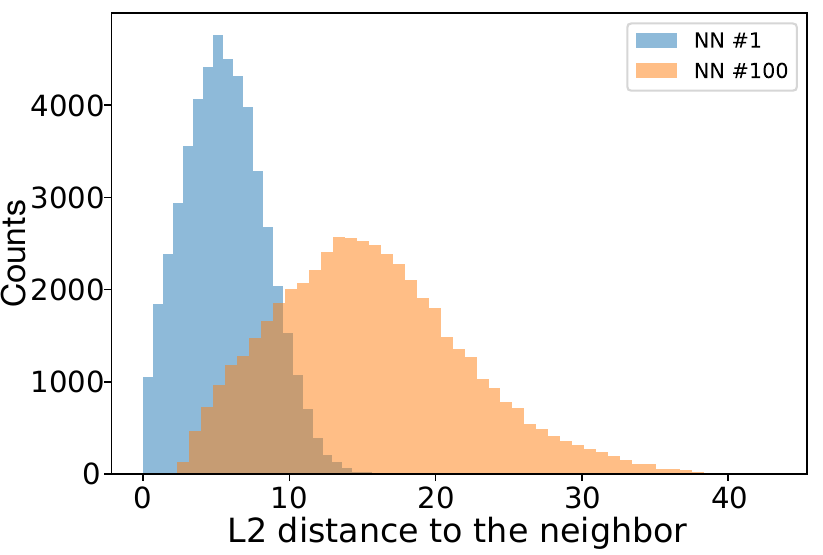} 
          \subcaption{}
    \end{minipage}
    \hspace*{15pt}
    \begin{minipage}[t]{0.4\linewidth}          \includegraphics[width=0.95\linewidth]{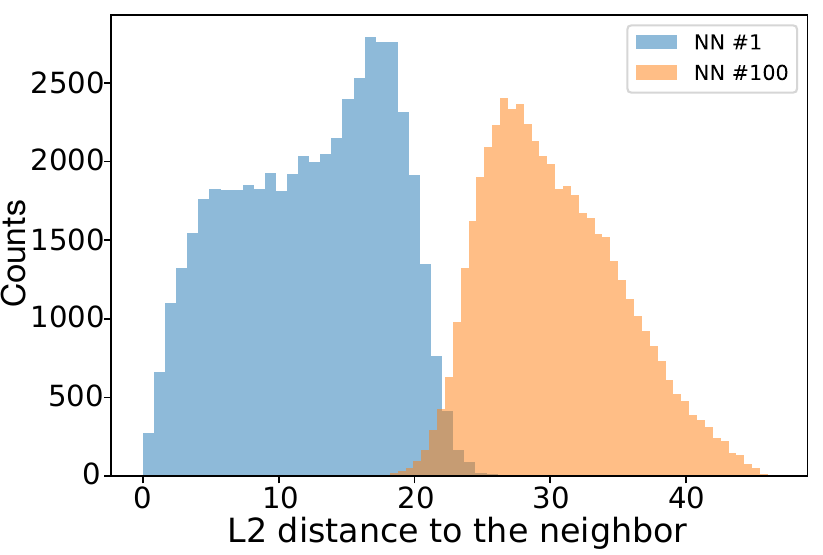}
          \subcaption{}
    \end{minipage}
    \vspace*{-5pt}
    \caption{Histograms of distances from a query point to its nearest neighbor (blue) and $100^\textrm{th}$ nearest neighbor (orange) for (a) VICReg and (b) VICReg + {\emc}.
    For VICReg, the histograms have significant overlap indicating less separability between points in embedding space. We show that continued pre-training with {\emc} reduces the overlap and ensures greater separability between points for downstream tasks.}
    \label{fig:hist}
    \vspace{-5pt}
\end{figure*}

\subsection{Ablation study: Embedding separability under different criteria}
One of the advantages of a maximum-entropy embedding is that data is well separated in the embedding space thereby preserving discriminability for downstream tasks. Inspired by \cite{spreadingvectors}, in \Cref{fig:hist}, we show the histogram of distances to the nearest and $100^{\textrm{th}}$ nearest neighbors for all points in the ImageNet validation set, computed using VICReg embeddings before (left) and after (right) continued pre-training using our maximum-entropy criterion. We note that for VICReg, the two histograms have significant overlap signifying that the $100^{\textrm{th}}$ nearest neighbor for a point is often closer than the $1^{\textrm{st}}$ nearest neighbor for another point, suggesting low separability that affects performance on downstream task. Continued pre-training using our criterion significantly reduces this overlap ensuring greater discriminability for downstream tasks.

\subsection{Comparative analysis: Number of continued pre-training Epochs}

\begin{figure}[h]
    \centering
    \vspace{-8pt}
    \includegraphics[width=0.9\linewidth]{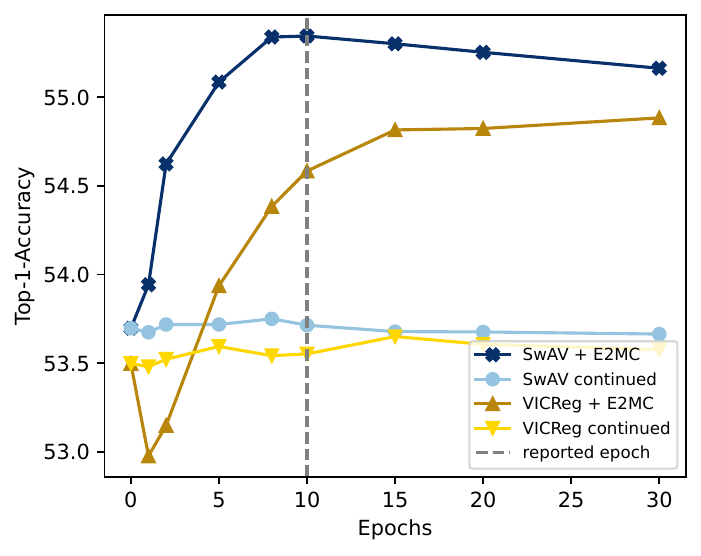}
    \vspace*{-10pt}
    \caption{Top-1-Accuracy of a linear classifier trained using 1\% ImageNet labels at different epochs of continued pre-training. Continued pre-training with our criteria (\emc) outperforms other baselines, and performance beyond the reported ten epochs either improves marginally or degrades depending on the method.}
    \label{fig:ftepochs}
    \vspace{-5pt}
\end{figure}

In this experiment, we show results from intermediate epochs of continued pre-training for a total of $30$ epochs (i.e., $3\times$ longer training than the reported results in \mbox{\Cref{tab:results}}).
In \Cref{fig:ftepochs}, we show the performance of SwAV (800 epochs) and VICReg (1000 epochs) base models at epoch 0, and how they evolve during continued pre-training with our {\emc} criterion compared to their base criterion.
It is evident that simply training longer using the base criterion (continued) provides no added benefit. We also notice a rapid improvement in performance until epoch $10$ after which the performance only improves marginally (e.g., for VICReg) or starts degrading (e.g., for SwAV), and therefore recommend $10$ epochs of continued pre-training as a good trade-off between performance and training time.

\section{DISCUSSION}
\label{sec:discussion}

ResNet-50 models pre-trained with SSL methods are the workhorse of computer vision applications across industry.
Despite significant efforts and resources devoted to the development of effective pre-training methods, 
the downstream performance of pre-trained ResNet-50 models had plateaued.
We have presented a simple method to maximize the entropy of pre-trained embeddings that consistently improves the performance of already-highly-optimized publicly available pre-trained ResNet-50 models with only a handful of training epochs.
In downstream tasks where only a small number of training samples is available, our method can be used to leverage the full potential of state-of-the-art models with rapid continued pretraining on as few as one GPU within a day. 
We also show that other methods for maximizing entropy do not converge as quickly as our method in this challenging setting, likely because they rely on high dimensional statistics or make fundamental assumptions about the underlying data distribution.
Moreover, by showing improvements on a variety of SSL methods, we hope that our general-purpose criterion can be extended to larger transformer-based models in the future, and squeeze any remaining gains out of them.

\section{CONCLUSION}
\label{sec:conclusion}
We proposed a simple add-on criterion for self-supervised learning motivated by information-theoretic principles and applicable to a wide variety of SSL methods.
We demonstrated empirically that the proposed criterion has desirable properties and that---with only a handful of epochs of continued pre-training---it is possible to achieve consistent and, in some cases, significant improvements in downstream-task performance across a selection of computer vision tasks.\vspace*{-20pt}

\clearpage

\clearpage

\section*{Acknowledgments}

The authors would like to thank Dmitry Petrov, Fabien Delattre, and Edmond Cunningham for helpful discussions and feedback. We would also like to thank the anonymous reviewers for their invaluable comments and suggestions to improve our work.
Part of this work utilized resources from Unity, a collaborative, multi-institutional high-performance computing cluster managed by UMass Amherst Research Computing and Data.

\bibliography{references}
\bibliographystyle{plainnat}
\begin{appendices}

\crefalias{section}{appsec}
\crefalias{subsection}{appsec}
\crefalias{subsubsection}{appsec}

\setcounter{equation}{0}
\setcounter{figure}{0}
\setcounter{table}{0}
\renewcommand{\theequation}{\thesection.\arabic{equation}}
\renewcommand\thefigure{\thesection.\arabic{figure}}
\renewcommand{\thetable}{\thesection.\arabic{table}}
\onecolumn

\section*{\LARGE Appendix}

\section{Algorithm for Continued Pre-Training with {\emc}}
\label{app:algo}
\begin{algorithm}[h!]
  \caption{PyTorch pseudocode for our max-entropy augmentation criterion}
  \label{alg:method}
    \definecolor{codeblue}{rgb}{0.25,0.5,0.5}
    \definecolor{codekw}{rgb}{0.85, 0.18, 0.50}
    \newcommand{\algofontsize}{11.0pt}
    \lstset{
      backgroundcolor=\color{white},
      basicstyle=\fontsize{\algofontsize}{\algofontsize}\ttfamily\selectfont,
      columns=fullflexible,
      breaklines=true,
      captionpos=b,
      commentstyle=\fontsize{\algofontsize}{\algofontsize}\color{codeblue},
      keywordstyle=\fontsize{\algofontsize}{\algofontsize}\color{black},
      tabsize=4,
    }
\vspace{-3pt}
\begin{lstlisting}[language=python]
# sample training loop
for x in loader:
	# two random views of x
	x_a, x_b = augment(x)

	# compute embeddings
	z_a = f(x_a) # N x D
	z_b = f(x_b) # N x D

	# base SSL loss
	base_loss = ssl_loss(z_a, z_b)
    
	# our criterion
	ent_z_a, cov_z_a = max_ent_criterion(z_a)
	ent_z_b, cov_z_b = max_ent_criterion(z_b)
	ent_loss = ent_z_a + ent_z_b
	cov_loss = cov_z_a + cov_z_b

	# final loss
	loss = base_loss - beta * ent_loss + gamma * cov_loss

	# optimization step
	loss.backward()
	optimizer.step()

def max_ent_criterion(x, type):
	if type == 'hypercube':
		# apply the sigmoid transformation
		x_hyper = torch.sigmoid(x)
	elif type == 'hypersphere':
		# apply the  CDF of 0-mean, 1 variance gaussian
		x_hyper = 0.5 * (1 + torch.erf(x / math.sqrt(2)))
	ent_loss = m_spacings_estimator(x_hyper)
	cov_loss = sample_cov_estimator(x_hyper)
	return (ent_loss, cov_loss)

def m_spacings_estimator(x):
	n = x.shape[0]  # batch size
	m = round(math.sqrt(n))  # window size
	eps = 1e-7  # small constant to avoid underflow
	x, _ = torch.sort(x, dim=0)  # order statistics
	x = x[m:] - x[:n - m]  # m-spaced differences
	x = x * (n + 1) / m
	marginal_ents = torch.log(x + eps).sum(dim=0) / (n - m)
	return marginal_ents.mean()

def sample_cov_estimator(x):
	n, d = x.shape
	x = x - x.mean(dim=0)  # mean subtraction
	cov_x = (x.T @ x) / (n - 1)  # sample covariance matrix
	cov_loss = off_diagonal(cov_x).pow(2).sum().div(d)
	return cov_loss
\end{lstlisting}
\vspace{-3pt}
\end{algorithm}
\vspace{-3pt}

\newpage
\section{Full Implementation Details}
\label{app:implement}

\subsection{Continued Pre-training}

Continued pre-training involves starting from a pre-trained checkpoint of a base SSL method and training for \emph{exactly} 10 epochs with an additional criteria.
This stage uses $2\times$NVIDIA Quadro RTX8000 GPUs with 48GB VRAM for continued pre-training of each model. Training times for 10 epochs of continued pre-training are 10 hrs for VICReg, 13 hrs for SwAV, and 14 hrs for SimSiam. Due to the extremely limited number of these GPUs, we could not train models from scratch, or fine-tune with bigger batch sizes, or run extensive grid searches for hyperparameters.
We will now detail the different criteria used and their associated hyperparameters for various base SSL methods.

\subsubsection{VICReg}
We start from the 1000-epoch checkpoint for VICReg \citep{vicreg} with ResNet-50 backbone, and 3-layer MLP (8192-8192-8192) as projector architecture. 
For continued pretraining, our criterion is applied to the final projector embeddings $Z$ mapped through a sigmoid transformation. 
We use the default coefficients for the VICReg loss function (\Cref{eq:vicregloss})  $\lambda = \mu = 25,\ \nu=1$, and coefficients used for our loss in Eqn. (\ref{eq:fullloss}) are $\beta=1000$, $\gamma=100$. 
We do continued pre-training for 10 epochs, with a learning rate of $0.003$ (i.e., $0.01\times$ the base learning rate used to train VICReg), batch size of $512$, and all other hyperparameters left unchanged from the original method. The same settings are used for continued training ablation using the base loss only. 

\subsubsection{SwAV}
We run experiments using both 400-epoch and 800-epoch checkpoints released for SwAV \citep{swav} with multicrop, using resnet50-backbone and 2-layer MLP (2048-128) as projector architecture. 
For continued pretraining, our criterion is applied to the projector embeddings $Z$ before the cluster assignment layer and before normalization after mapping through the CDF function.
We use default hyperparameters for SwAV loss (\Cref{eq:swavloss}) namely $\tau=0.1$, and apply our loss only to the embeddings of full-resolution crops (two views) and not the low resolution multiple crops for computational efficiency, with coefficients $\beta=1$ and $\gamma=25$. See \Cref{app:coeff} for an ablation study on the coefficients.
We do continued pre-training for 10 epochs, with a learning rate of $0.001$ (i.e., $0.01\times$ the base learning rate used to train SwAV), batch size of $512$, and all other hyperparameters left unchanged from the original method. The same settings are used for continued training ablation without our loss. 

We will now describe the ablation studies that train SwAV models further using an alternative criterion and associated hyperparameters.

\paragraph{Variance-Covariance Regularization \citep{vicreg}.} We use the variance and covariance regularization losses from the VICReg objective in \Cref{eq:vicregloss}, to minimize the following loss
\begin{align}
    \mathcal{L(\theta)} 
    = 
    \mathcal{L}^{\mathrm{SSL}}(\theta) + \mu \mathcal{L}^{\mathrm{Variance}}(\theta) + \nu \mathcal{L}^{\mathrm{Covariance}}(\theta) \label{eq:vcreg}
\end{align}
We set $\mu=0.1$ and $\nu=0.001$ and this loss is applied only on the embeddings of the two full resolution crops to be consistent with our setting. These hyperparameters were determined experimentally by searching over $\mu=[0.01, 0.1, 1, 25]$ and $\nu=[0.001, 0.005, 0.01, 0.1, 1, 25]$ on the validation set using linear classifier trained on 1\% ImageNet labels. 

\paragraph{Uniformity loss from AUH \citep{auh}.} We use the following loss: 
\begin{align}
    \mathcal{L(\theta)} 
    = 
    \mathcal{L}^{\mathrm{SSL}}(\theta) + \lambda \mathcal{L}^{\mathrm{Uniform}}(\theta) \label{eq:unifloss}
\end{align}
where,
\begin{align}
\begin{split}
    \mathcal{L}^{\mathrm{Uniform}}(\theta) \triangleq 
    & 
    \log~\mathbb{E}_{[Z^p, Z^q \iidsim p(z)]} \Big[ G_t(Z_{\theta}^p, Z_{\theta}^q) \Big]
    \\
    &
    + 
    \log~\mathbb{E}_{[Z^{'p}, Z^{'q} \iidsim p(z')]} \Big[ G_t(Z_{\theta}^{'p}, Z_{\theta}^{'q}) \Big], \quad t>0
\end{split}
\end{align}
where $p, q \in \{1,\cdots, n\}$, and $G_t$ is defined as the average pairwise gaussian potential between two embedding vectors:
\begin{align}
    G_t(Z_{\theta}^p, Z_{\theta}^q) = \exp\left({-t \Vert{Z^p - Z^q}\Vert_2^2}\right)
\end{align}
In practice this is applied to all embedding pairs from each of the full resolution crops (each of the two views) to be consistent with our setting. The hyperparameters used are $\lambda=0.5$ and $t=2$, and continued pre-training is done for 10 epochs with the same training hyperparameters. The hyperparameters for this method are guided by the original paper and experimentally verified on $1\%$-ImageNet split as before. 

\paragraph{Maximum Manifold Capacity Representations (MMCR; \cite{mmcr})} Consider multiview embeddings $Z^{(v)} \in \mathbb{R}^{d\times k}$ for input views $v\in \{1, \cdots, V\}$. The centroid embedding $C$ is then considered an average embedding across the views $C = \frac{1}{V} \sum\nolimits_{v=1}^V Z^{(v)}$. 
We use the following loss
\begin{align}
    \mathcal{L(\theta)} 
    &= 
    \mathcal{L}^{\mathrm{SSL}}(\theta) - \lambda \Vert C \Vert_{*} \\
    &\textrm{where} \quad \Vert C \Vert_{*} = \sum\nolimits_{r=1}^{rank(C)} \sigma_r (C), \nonumber
\end{align}
where $\Vert \cdot \Vert_{*}$ is defined as the nuclear norm of a matrix, and $\sigma_r (C)$ is the $r$-th singular value of C.
This loss is applied over all views, i.e., multi-resolution crops of SwAV with coefficient $\lambda = 0.005$, and $V=8$. The coefficient is determined by grid search over $\lambda = [0.001, 0.005, 0.1, 0.5, 1, 2]$ verified using the same 1\%-ImageNet split as before.

\subsubsection{SimSiam}
We start from the 100-epoch checkpoints released for SimSiam  \citep{simsiam}, using resnet50-backbone and 3-layer MLP (2048-2048-2048) as projector, and 2-layer MLP (2048-512) as predictor architecture. 
For continued pretraining, our criterion is applied to the projector embeddings $Z$ on both branches before the predictor, mapped through a CDF function. 
The projector embeddings thus serve as uniformly distributed points on the hypersphere that the predictor has to map to.
We use default coefficients for SimSiam loss, and apply our loss with coefficients $\beta=0.001$ and $\gamma=0.01$. We do continued pre-training for 10 epochs, with a learning rate of $0.001$ (i.e., $0.01\times$ the base learning rate used to train SimSiam), batch size of $512$, and all other hyperparameters left unchanged from the original method. The same settings are used for continued training ablation without our loss.

\subsection{Evaluation}
In this stage, the final representations from ResNet-50 backbone are used to train classifiers on different datasets in order to evaluate the quality of the representations. Training hardware includes $4\times$NVIDIA RTX 2080TI GPUs with 11GB VRAM for each training.

\subsubsection{Linear evaluation on ImageNet.}
Following standard practice, we train linear classifiers using frozen ResNet-50 representations on 1\% (12,811 images), 10\% (128,117 images), and 100\% (1,281,176 images) of ImageNet labels (using predefined splits from \citep{simclr}) for 100 epochs, and report the top-1 accuracy on the validation set containing 50,000 images and 1,000 classes.

For VICReg, we use the SGD optimizer with learning rate 0.02 and cosine decay, batch size of 256, and a weight decay of $10^{-4}$ for $1\%$ and $10\%$ splits, and $10^{-6}$ for $100\%$ split respectively.

For SwAV, we use the SGD optimizer with learning rate 0.3 and cosine decay, batch size of 256, and a weight decay of $10^{-6}$ for all splits.

For Simsiam, we use the LARS optimizer with weight decay 0 and cosine decay for learning rate as follows. For the $1\%$ split, we use learning rate 2.0 and batch size 256. For the $10\%$ split, we use learning rate 0.2 and batch size 2048. For the $100\%$ split, we use learning rate 0.1 and batch size 2048. 

\subsubsection{Semi-supervised learning on ImageNet.}
We perform semi-supervised evaluation by finetuning the whole backbone with a classifier on a subset of available labels. 

For VICReg, we use the SGD optimizer with batch size 256, cosine learning rate schedule, and no weight decay, and train for 20 epochs using learning rate 0.03 for the backbone and 0.08 for the linear classifier in the 1\% labels setting, and learning rate 0.01 for the encoder and 0.1 for the linear classifier in the 10\% labels setting. Unfortunately, we're unable to reproduce the numbers reported in the paper (54.8\% and 69.5\% respectively) exactly using these prescribed settings, and report our closest reproduced values in \Cref{tab:results}. 

For SwAV, use the SGD optimizer with batch size 256, step decay of 0.2 at epochs 12 and 16 for a total of 20 epochs, and no weight decay using learning rate 0.02 for the backbone and 5 for the linear classifier in the 1\% labels setting, and learning rate 0.01 for the encoder and 0.2 for the linear classifier in the 10\% labels setting.

For SimSiam, the semi-supervised experiments were not conducted in the original paper, and therefore we skip this in our experiments.

\subsubsection{Transfer learning performance on other datasets.} 
Following \citep{pirl, vissl}, we show how representations updated using our method on ImageNet dataset generalize to downstream linear classification on other datasets such as iNaturalist 2018 \citep{inaturalist} and Pascal VOC 2007 \citep{pascal}. 

For iNat18 (437,513 images and 8,142 classes), we use \verb|res5| features from the ResNet-50 backbone (before average pooling layer) subsampled to 8192-d using an average pooling layer of size (6, 6) and stride 1, followed by a batch normalization layer. 
A linear classifier is then trained on top of these representations using the SGD optimizer with batch size 256, weight decay $10^{-4}$, momentum 0.9, and learning rate 0.01 reduced by a factor of 10 at epochs 24, 48, and 72, for a total of 84 epochs. 
These hyperparameters are used consistently across all methods, and we find that for SwAV, we obtain better performance (49.72) than reported in the original paper (48.6). 

For VOC07 (5,011 images and 20 classes), we train linear SVMs on top of final average pooled representations (2048-d) from ResNet-50 backbone using the VISSL library \citep{vissl} and report the mean Average Precision (mAP) of multi-label object classification on the validation set. In this setting, we were unable to reproduce numbers reported in the papers exactly due to missing hyperparameter details and default values in the library not working well. The numbers reported in the paper are using the following $C$ values: 
$[0.000001, 0.000003, 0.00001, 0.00003, 0.0001, 0.0003, 0.001, 0.003, 0.01, 0.03, 0.1, 0.3, 1, 2, 5, 10, \\15, 20, 50, 100, 200, 500, 1000]$. We get close to the performance reported in original papers, and mark our results where there are significant differences, such as SwAV (88.56 vs. the reported 88.9).

\section{Maximum Entropy Augmentation Criteria: Further Details}
\label{app:components}

\subsection{Sample Covariance}
We estimate the squared off-diagonal covariance matrix entries using the sample covariance. 

Let ${\check{\bar{Z}}}_{\theta} = \check{Z}_{\theta} - \frac{1}{n} \sum\nolimits_{i = 1}^{n} \check{Z}_{\theta}^{i} $ for $\check{\bar{Z}}_{\theta}, \check{Z}_{\theta} \in \mathbb{R}^{n \times d}$.
Now, consider the sample covariance estimator
\begin{align}
    \widehat{\mathrm{Cov}}_{jk}(\check{Z}_{\theta}^{1}, ..., \check{Z}_{\theta}^{n})
    =
    \frac{1}{n-1} \sum\nolimits_{i=1}^{n} \check{\bar{Z}}_{\theta j}^{i} \check{\bar{Z}}_{\theta k}^{i}
\end{align}
for the $j$th and $k$th dimension of $\check{Z}_{\theta}$.

Letting $K_{\theta} = \check{\bar{Z}}_{\theta}^\top \check{\bar{Z}}_{\theta}$, we define
\begin{align}
    \mathcal{L}^{\mathrm{Covariance}}(\theta)
    =
    \frac{1}{nd} \left\|(K_{\theta}-\operatorname{diag}(K_{\theta}))\right\|_F^2=\frac{1}{d} \sum\nolimits_{j=1,k=1}^d \mathbb{I}[ k\neq j ] \widehat{\mathrm{Cov}}_{jk}(\check{Z}_{\theta}^{1}, ..., \check{Z}_{\theta}^{n})^2 ,
    \label{eq:sq_cov_estimator}
\end{align}
where $\| \cdot \|_{F}^{2}$ is the squared Frobenius norm.

\section{Further Empirical Results}

\subsection{Statistical Significance of Our Results}
\label{app:statsig}

\begin{table}[h]
\caption{Results for statistical tests conducted on the ImageNet validation set using both linear evaluation and semi-supervised learning protocol. The results show the test statistic for Paired Permutation test and McNemar's test along with associated $p$-values. Improvements achieved by {\emc} over the baseline are statistically significant.}
\label{tab:statsig}
\resizebox{\textwidth}{!}{%
\begin{tabular}{@{}lccccccc@{}}
\toprule
\begin{tabular}[c]{@{}l@{}}Evaluation \\ protocol\end{tabular} &
  \begin{tabular}[c]{@{}l@{}}VICReg \\ (1000 epochs)\end{tabular} &
  \begin{tabular}[c]{@{}l@{}}VICReg + {\emc}\\ (1010 epochs)\end{tabular} &
  \begin{tabular}[c]{@{}l@{}}Observed Mean \\ $\Delta$ ($\mu_{\textrm{\emc}}$ - $\mu_{\textrm{base}}$)\end{tabular} &
  \begin{tabular}[c]{@{}l@{}}Permutation \\ test $p$-value\end{tabular} &
  \begin{tabular}[c]{@{}l@{}}Effect size \\ (Cohen’s $d$)\end{tabular} &
  \begin{tabular}[c]{@{}l@{}}McNemar’s test  \\ $\chi^2 = \frac{(n_{01} - n_{10})^2}{n_{01} + n_{10}}$\end{tabular} &
  \begin{tabular}[c]{@{}l@{}}McNemar’s \\ $p$-value\end{tabular} \\ \midrule
Linear, 1\% labels    & 53.50 ± 0.11 & 54.54 ± 0.05 & 0.010 & < 0.001 & 0.05 & 108.79 & < 0.001 \\
Linear, 10\% labels   & 66.57 ± 0.02 & 66.82 ± 0.05 & 0.002 & \hspace{0.3em} 0.02            & 0.01 & 5.25   & \hspace{0.3em} 0.02            \\
Linear, 100\% labels  & \hspace{-3.6em} 73.20        & 73.45 ± 0.07 & 0.002 & \hspace{0.8em} 0.003           & 0.01 & 8.46   & \hspace{0.8em} 0.003           \\
Semi-sup, 1\% labels  & 54.53 ± 0.12 & 55.05 ± 0.08 & 0.006 & < 0.001 & 0.03 & 47.38  & < 0.001 \\
Semi-sup, 10\% labels & 67.97 ± 0.03 & 68.12 ± 0.04 & 0.002 & < 0.001 & 0.02 & 12.12  & < 0.001 \\ \bottomrule
\end{tabular}%
}
\end{table}

In this section, we evaluate the statistical significance of the improvements obtained by our method {\emc} over the VICReg (1000 epochs) baseline, as reported in \Cref{tab:results}. To do this, we conduct a paired permutation test \citep{good} and McNemar's test \citep{mcnemar} on the ImageNet validation set, consisting of 50k samples. These tests were chosen to analyze differences in correctness of predictions between the two models, and prediction disagreements respectively, as we have paired data for both models. For running these tests, we picked a single model each for the baseline and for our method with performance closest to the reported mean accuracy on the ImageNet 1\%-labels setting. 

The $p$-values for both paired permutation test and McNemar’s test (\Cref{tab:statsig}) confirm that improvements on ImageNet obtained by our method under all settings are statistically significant, as we reject the null hypothesis with $p$-values well under 0.05 (the chosen significance level). Note that while the effect size (Cohen’s $d$) in all these tests is small, our results are still impactful because achieving significant performance improvements on ImageNet with few epochs of continued pre-training is challenging, as most SSL models are already highly optimized.

\subsection{Ablation Study: Contribution of Different Components in Our Objective Criterion}
\label{app:coeff}

\setlength{\tabcolsep}{12pt}
\begin{table}[h]
\centering
\caption{Contribution of entropy and covariance components controlled by their coefficients $\beta$ and $\gamma$ resp.  We report Top-1-Accuracy (\%) of linear classifier trained on ImageNet, using SwAV + {\emc} method (810 epochs).}
\label{tab:coeff}
\begin{tabular}{@{}lllll@{}}
\toprule
$\beta$ & $\gamma$ & 1\% labels    & 10\% labels   & Notes                                              \\ \midrule
0       & 0        & 53.69         & 68.87         & Baseline (SwAV contd.) - no entropy, no covariance \\
1       & 0        & 53.85 ($\uparrow$ 0.16) & 68.81 ($\downarrow$ 0.06) & Only entropy                                       \\
0       & 1        & 53.63 ($\downarrow$ 0.06) & 68.83 ($\downarrow$ 0.04) & Only covariance                                    \\
1   & 25 & \textbf{55.27 ($\uparrow$ 1.58)} & \textbf{68.98 ($\uparrow$ 0.11)} & Reported (SwAV $+$ {\emc}) - entropy $<$ covariance \\
1       & 0.1      & 53.92 ($\uparrow$ 0.23) & 68.81 ($\downarrow$ 0.06) & Entropy $+$ some covariance                          \\
0.1     & 1        & 54.09 ($\uparrow$ 0.40) & 68.89 ($\uparrow$ 0.02) & Covariance $+$ some entropy                          \\
1       & 1        & 54.29 ($\uparrow$ 0.60) & 68.90 ($\uparrow$ 0.03) & Entropy $==$ Covariance                              \\
1       & 10       & 54.99 ($\uparrow$ 1.30) & 68.91 ($\uparrow$ 0.04) & Entropy $<$ Covariance                       \\
10      & 1        & 51.53 ($\downarrow$ 2.16) & 68.01 ($\downarrow$ 0.86) & Entropy $>$ Covariance                    \\
10      & 25       & 55.12 ($\uparrow$ 1.43) & 69.11 ($\uparrow$ 0.24) &                                                    \\
0.1 & 25 & 53.88 ($\uparrow$ 0.19)          & 68.86 ($\downarrow$ 0.01)          & Entropy $\ll$ Covariance                 \\
0       & 25       & 53.66 ($\downarrow$ 0.03) & 68.83 ($\downarrow$ 0.04) & No entropy, high covariance                        \\ \bottomrule
\end{tabular}%
\end{table}

Here, we analyze the top-1 accuracy on the ImageNet validation set (using linear probes trained on 1\% and 10\% of the training labels) for continued pre-training of SwAV (800 epochs) using our proposed objective, albeit with different coefficients for the marginal entropy component and pairwise covariance component ($\beta$ and $\gamma$ respectively in \Cref{eq:fullloss}). This shows the effect of removing or \textit{reducing} the contribution of either of the two components (marginal entropy and covariance) in our proposed objective.

Our hypothesis is that for {\emc} to work as intended, we need both of these components in the loss to avoid learning degenerate embedding distributions. The results shown in \Cref{tab:coeff} confirm that removing either one of these components ($\beta=1, \gamma=0$ or $\beta=0, \gamma=1$) either hurts performance compared to the baseline ($\beta=0, \gamma=0$), or achieves smaller performance improvements compared to the best possible model ($\beta=1, \gamma=25$). Moreover, these results also show how performance changes as a function of varying contributions from the entropy and covariance losses, leading us to conclude that our reported results are using near-optimal hyperparameters.

\subsection{Sample Distributions over Two Dimensional Marginals}

\begin{figure}[t]
    \centering
    \begin{minipage}{0.46\linewidth}
        \centering
        \includegraphics[width=\linewidth]{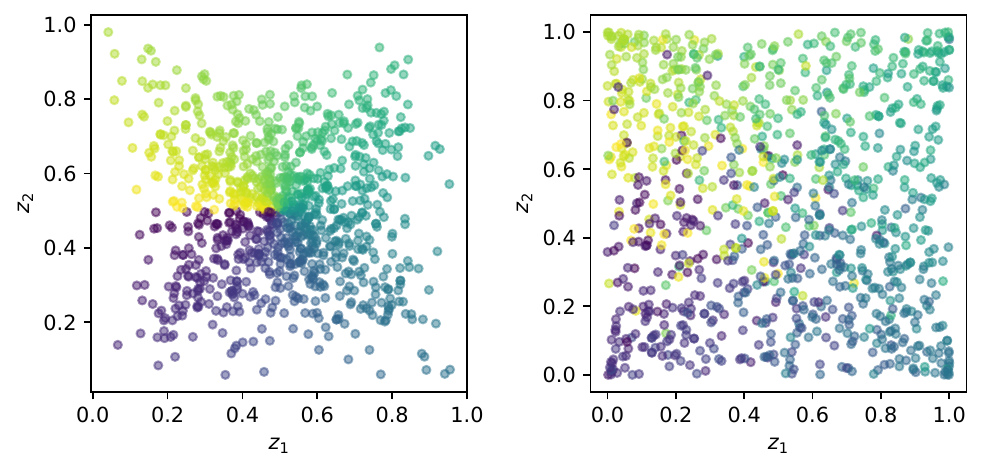}
        \subcaption{}
        \label{fig:subfig_a}
    \end{minipage}
    \hfill
    \begin{minipage}{0.46\linewidth}
        \centering
        \includegraphics[width=\linewidth]{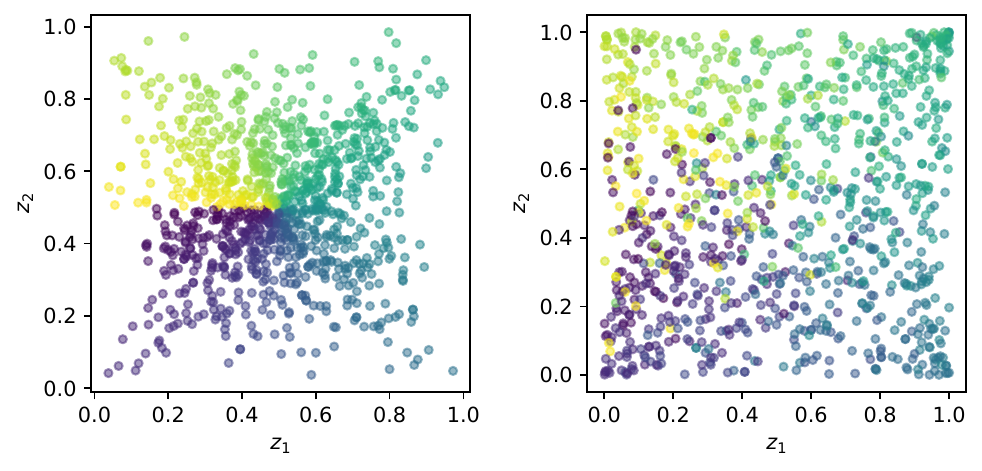}
        \subcaption{}
        \label{fig:subfig_b}
    \end{minipage}

    \vspace{0.5em}

    \begin{minipage}{0.46\linewidth}
        \centering
        \includegraphics[width=\linewidth]{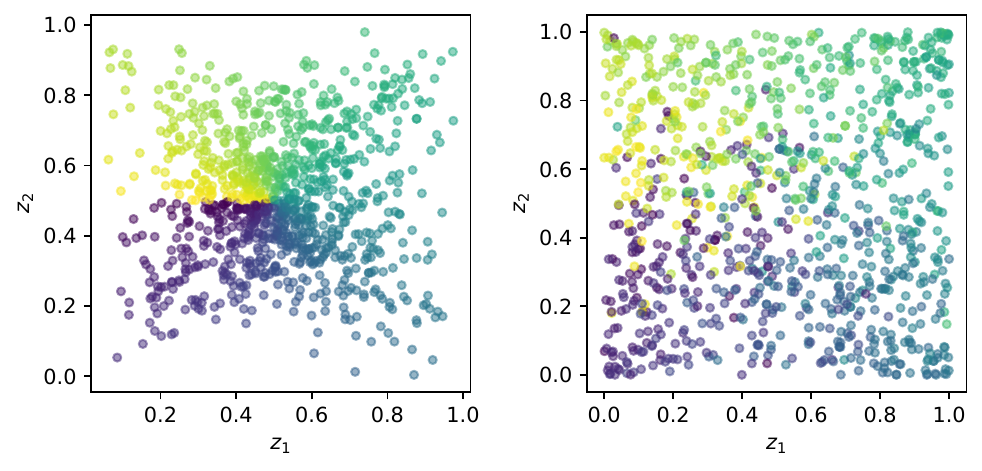}
        \subcaption{}
        \label{fig:subfig_c}
    \end{minipage}
    \hfill
    \begin{minipage}{0.46\linewidth}
        \centering
        \includegraphics[width=\linewidth]{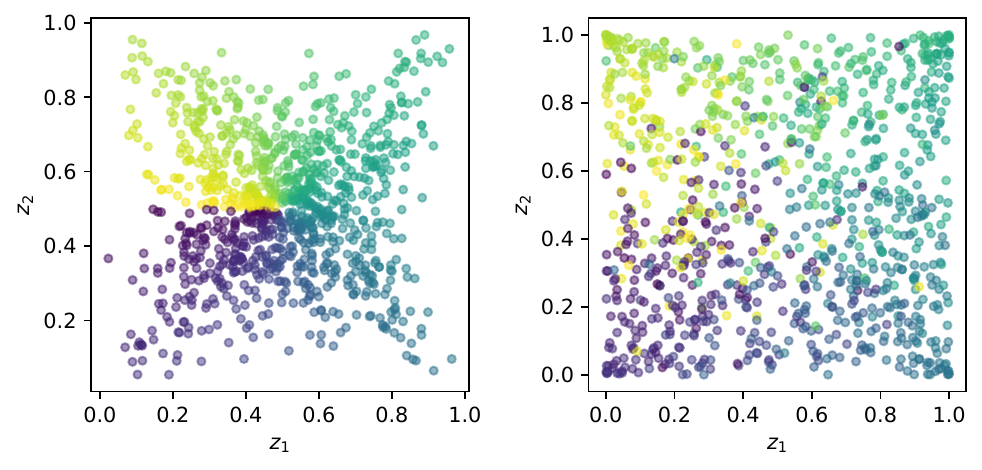}
        \subcaption{}
        \label{fig:subfig_d}
    \end{minipage}

    \vspace{0.5em}

    \begin{minipage}{0.46\linewidth}
        \centering
        \includegraphics[width=\linewidth]{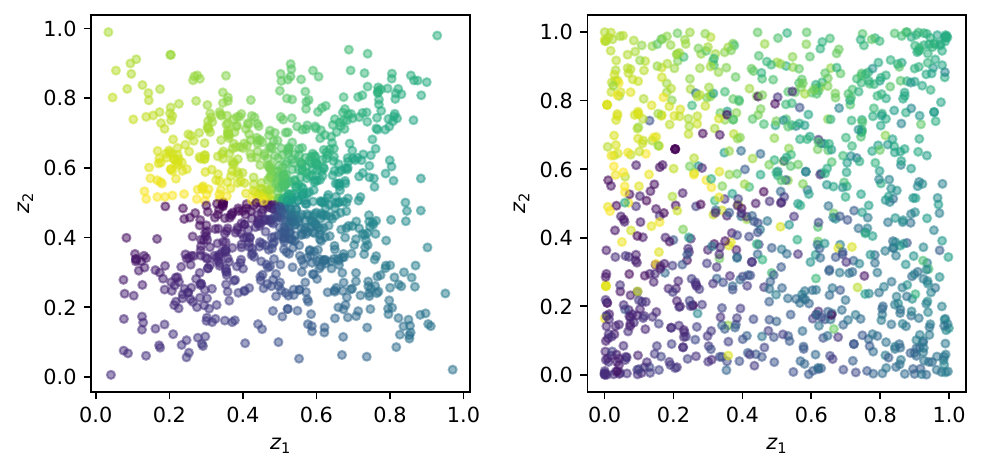}
        \subcaption{}
        \label{fig:subfig_e}
    \end{minipage}
    \hfill
    \begin{minipage}{0.46\linewidth}
        \centering
        \includegraphics[width=\linewidth]{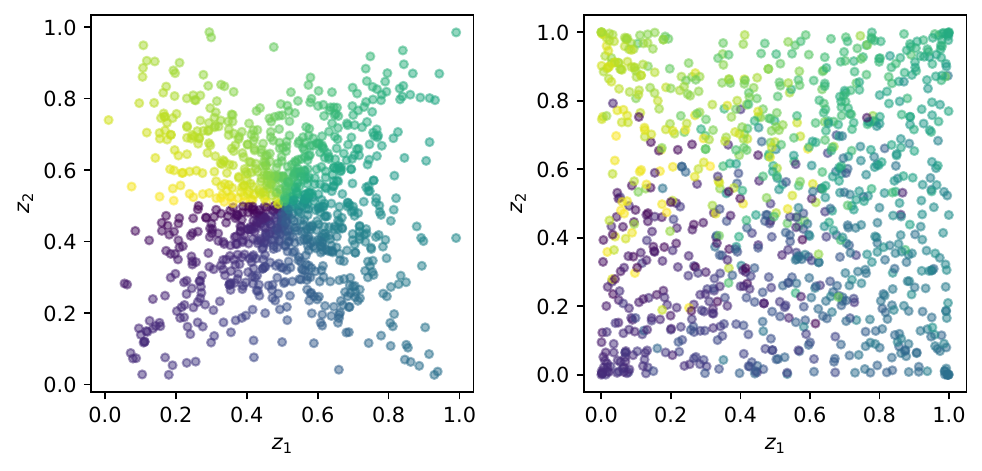}
        \subcaption{}
        \label{fig:subfig_f}
    \end{minipage}

    \vspace{0.5em}

    \begin{minipage}{0.46\linewidth}
        \centering
        \includegraphics[width=\linewidth]{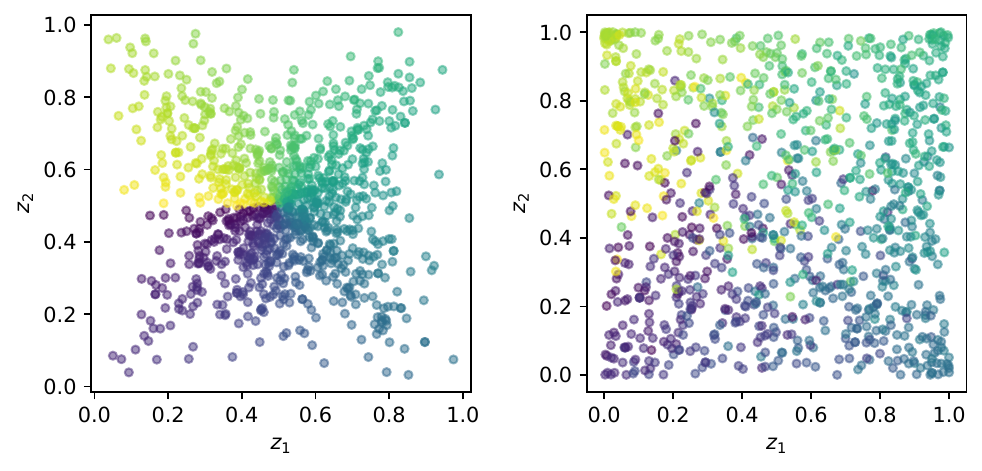}
        \subcaption{}
        \label{fig:subfig_g}
    \end{minipage}
    \hfill
    \begin{minipage}{0.46\linewidth}
        \centering
        \includegraphics[width=\linewidth]{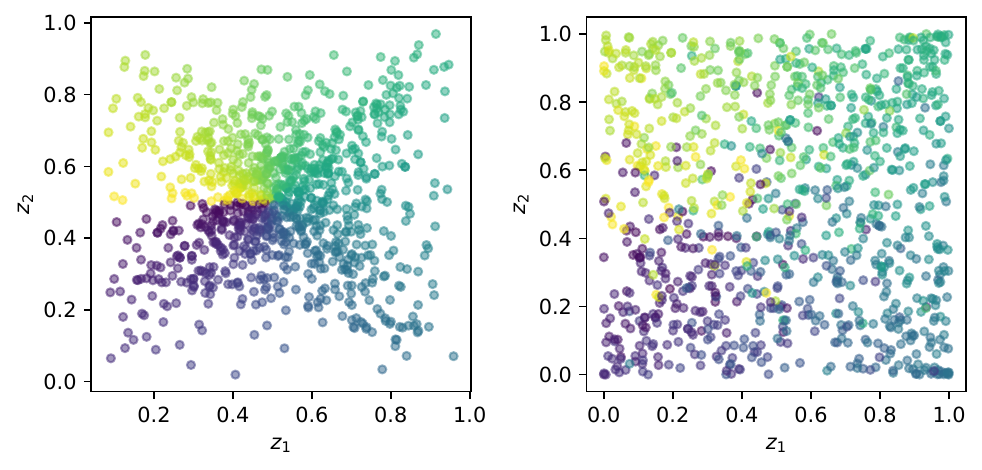}
        \subcaption{}
        \label{fig:subfig_h}
    \end{minipage}

    \vspace{0.5em}

    \begin{minipage}{0.46\linewidth}
        \centering
        \includegraphics[width=\linewidth]{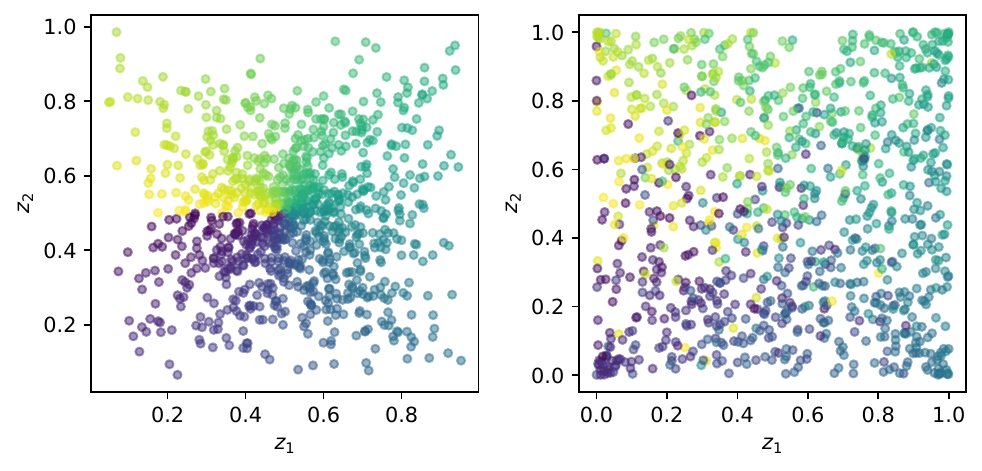}
        \subcaption{}
        \label{fig:subfig_i}
    \end{minipage}
    \hfill
    \begin{minipage}{0.46\linewidth}
        \centering
        \includegraphics[width=\linewidth]{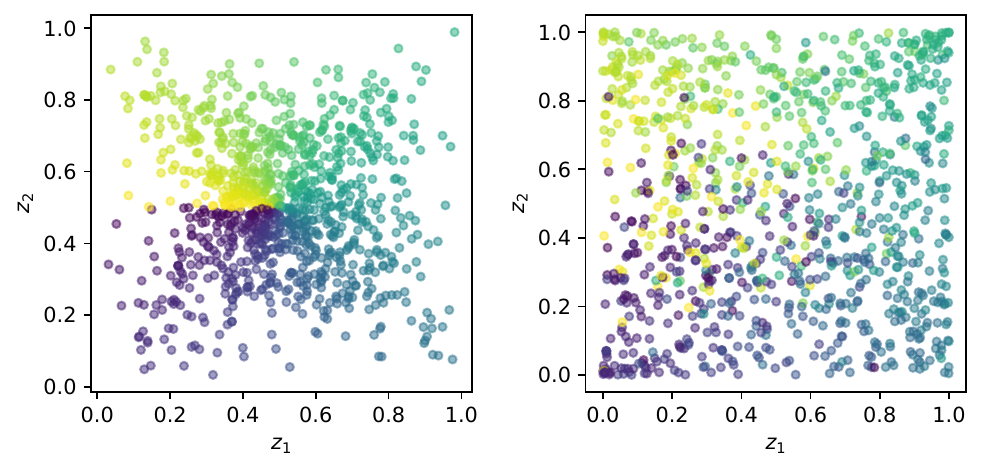}
        \subcaption{}
        \label{fig:subfig_j}
    \end{minipage}
    \caption{Plots (a)-(j) show sample distributions from VICReg (left) and VICReg + {\emc} (right) over a random pair of compact embedding dimensions, for a fixed set of data points.}
    \label{fig:moredists}
\end{figure}

\Cref{fig:moredists} shows more examples of sample distributions over a randomly selected pair of embedding dimensions (marginals) from VICReg \textit{before} and \textit{after} continued pre-training with our maximum-entropy criterion. Note how our method virtually always produces uniformly distributed marginals over any random pair even though this is not explicitly enforced by our loss. A fixed set of colors was assigned to the data points when plotting the ``before'' embeddings, and one can follow how the points were distributed by the application of our criterion by noting the relative distribution of colors in the ``after'' embeddings.

\end{appendices}

\end{document}